\documentclass[notitlepage,aps,pre,superscriptaddress]{revtex4-1}

\usepackage[utf8]{inputenc}
\usepackage[T1]{fontenc}
\usepackage{url}
\usepackage{hyperref}
\usepackage{thmtools,thm-restate}
\usepackage{enumitem}
\usepackage{sidecap}
\usepackage{adjustbox} 

\usepackage{natbib}
\usepackage{graphicx}
\usepackage{dcolumn}
\usepackage{amsmath}
\usepackage{comment} 
\usepackage{amssymb}
\usepackage{mathrsfs} 
\usepackage[normalem]{ulem}
\usepackage{bbm,bm}
\usepackage{subfigure}

\usepackage[dvipsnames]{xcolor}

\usepackage{mathtools}
\renewcommand{\doteq}{\vcentcolon=}

\usepackage{cancel}
\usepackage{rotating}
\usepackage{hyphenat}
\usepackage{algorithm}
\usepackage{algpseudocode}
\usepackage{multirow}
\usepackage{amsthm}
\theoremstyle{definition}

\theoremstyle{definition}

\newtheorem*{exercise*}{Exercise}
\theoremstyle{remark}

\begin{document}

 \nocite{behjoo_github}
	
	\title{Harmonic Path Integral Diffusion}

	\author{{\bf Hamidreza Behjoo} 
		\& {\bf Michael (Misha) Chertkov} 
		\\
		Program in Applied Mathematics \& Department of Mathematics, \\
         University of Arizona, Tucson, AZ 85721, USA;\\ 
	}

	\begin{abstract}	
\textbf{Harmonic Path Integral Diffusion} (H-PID) introduces a novel approach to sampling from complex, continuous probability distributions by creating a time-dependent “bridge” from an initial point to the target distribution. Formulated as a Stochastic Optimal Control problem, H-PID balances control effort and accuracy through a unique three-level integrable structure: \textbf{Top Level:} Potential, force, and gauge terms combine to form a linearly solvable Path Integral Control system based on Green functions. \textbf{Mid Level:} With quadratic potentials and affine force/gauge terms, the Green functions reduce to Gaussian forms, mirroring quantum harmonic oscillators in imaginary time. \textbf{Bottom Level:} For a uniform quadratic case, the optimal drift/control reduces to a convolution of the target distribution with a Gaussian kernel, enabling efficient sampling. Implementation-wise the low-level H-PID operates without neural networks, allowing it to run efficiently on standard CPUs while achieving high precision. Validated on Gaussian mixtures and CIFAR-10 images, H-PID reveals a “weighted state” parameter as an order parameter in a dynamic phase transition, signaling early completion of the sampling process. This feature positions H-PID as a strong alternative to traditional methods sampling, such as simulated annealing, particularly for applications that demand analytical control, computational efficiency, and scalability.
	\end{abstract}
	
	\maketitle

\section*{Extended Abstract}

In this manuscript, we present a novel approach for sampling from a continuous multivariate probability distribution, which may either be explicitly known (up to a normalization factor) or represented via empirical samples. Our method constructs a time-dependent bridge from a delta function centered at the origin of the state space at $t=0$, optimally transforming it into the target distribution at $t=1$. We formulate this as a \textbf{Stochastic Optimal Control} problem of the \textbf{Path Integral Control} type, with a cost function comprising (in its basic form) a quadratic control term, a quadratic state term, and a terminal constraint. This framework, which we refer to as \textbf{Harmonic Path Integral Diffusion} (H-PID), leverages an analytical solution through a mapping to an auxiliary quantum harmonic oscillator in imaginary time. The analytic expression for the optimal control function takes the form ${\bm u}^{(*)}(t)= a(t){\bm x}(t)-b(t)\hat{\bm x}(t;{\bm x}(t))$, where ${\bm x}(t)$ is the current state at time $t \in [0,1]$, and $\hat{\bm x}(t;{\bm x}(t))$ is the current weighted state, with the weight expressing marginal probability distribution of the optimal stochastic process and functions $a(t), b(t)$ and $\hat{\bm x}(t;{\bm x})$ found analytically via convolutions with the target distribution. 

The H-PID framework results in a set of efficient sampling algorithms, \textbf{without} the incorporation of \textbf{Neural Networks}. The algorithms are validated on two standard use cases: a mixture of Gaussians over a grid and images from CIFAR-10. The transparency of the method allows us to analyze the algorithms in detail, particularly revealing that the current weighted state is an {\bf order parameter} for the {\bf dynamic phase transition}, signaling earlier, at $t<1$, that the sample generation process is almost complete. We contrast these algorithms with other sampling methods, particularly simulated annealing and path integral sampling, highlighting their advantages in terms of analytical control, accuracy, and computational efficiency on benchmark problems.

Additionally, we extend the methodology to more general cases where the underlying stochastic differential equation includes an external deterministic, possibly non-conservative force, and where the cost function incorporates a gauge potential term. These extensions open up new possibilities for applying our framework to a broader range of statistics specific to applications.

	\section*{Why, What, How, and What Not of This Manuscript}
	
	{\bf \it Why:} This manuscript is motivated by the revolutionary developments in generative Artificial Intelligence (AI) over the past decade, particularly in Score-Based Diffusion (SBD) models of AI \cite{ho_denoising_2020,song_score-based_2021}. Of special relevance to our work is the breakthrough in generating images -- initially achieved in this domain and later extended to tools that rival Markov Chain Monte Carlo (MCMC) techniques for sampling from complex, multi-dimensional probability distributions \cite{midgley_flow_2023}. What is particularly noteworthy is that SBD models, as demonstrated by foundational contributions \cite{anderson_reverse-time_1982,sohl-dickstein_deep_2015}, have their roots in decades of research in physical applied mathematics, specifically in stochastic processes and non-equilibrium statistical mechanics. 
	{\bf \it What:} However, this manuscript is not focused on mainstream SBD models. Instead, we explore the development of alternative approaches with strong foundations in physical applied mathematics, via the so-called Schrödinger bridges, originating from Schrödinger's 1932 contributions \cite{schrodinger_sur_1932}. These approaches are also linked to engineering applied mathematics, particularly Stochastic Optimal Control (SOC) methods pioneered by Bellman \cite{bellman_dynamic_2003}. The Schrödinger bridges, developed as a stochastic generalization of optimal transport and connected to SOC \cite{chen_stochastic_2021,chen_optimal_2021}, have shown the potential to complement and enhance SBD frameworks \cite{zhang_path_2022,chen_likelihood_2023}.
	
	{\bf \it How:} Specifically, we build upon previous work that uses stochastic drift-diffusion models to generate complex probability distributions from simpler ones. In our case, the initial distribution is a delta function positioned at the origin of the state space. Following SOC-based designs for SBD, as in \cite{zhang_path_2022}, we formulate the process optimally. Moreover, this optimality leads to an analytic expression for the control/drift function, derived from the target distribution—either provided explicitly or through samples. This analytical formulation, which dramatically improves computational efficiency by replacing costly simulations with efficient computations, is made possible by the integrable structure of SOC formulations, known as Path Integral Control (PIC) \cite{kappen_path_2005,kappen_optimal_2012}. Further details and references are provided in Section \ref{sec:th-H}.
	
	This manuscript begins by developing the theoretical framework. Our main theoretical results are presented in Section \ref{sec:theory}, with the aim of transitioning into algorithms and applications. The process of moving from theory to algorithmic implementation and experiments is described in Section \ref{sec:th-to-alg}.

	{\bf \it What Not (yet):} It is important to note that the experiments we present here are standard but simple. They are aimed at understanding and meant to illustrate and test the algorithmic potential of our theoretical developments. However, much more work is needed to demonstrate that the resulting algorithms are competitive with the state-of-the-art in AI diffusion and related fields. See more on related recent developments and discussion of the path forward in Sections \ref{sec:recent},\ref{sec:conclusions}.
	
	Notably, the algorithms presented in this manuscript do not rely on Neural Networks (NN) \footnote{Similar to standard MCMC techniques, our sampling algorithm involves sequential operations and frequent conditional branching -- both nonlinear processes -- which makes it significantly more efficient to implement on a CPU than a GPU \cite{gelman_bayesian_2013}. This contrasts with NN-based algorithms, which benefit from the highly parallelized nature of GPUs due to their reliance on massive linear operations that are well-supported by GPU architecture.}. This makes our approach both transparent and explainable, in stark contrast to the 'black-box' nature of many NN-based methods. In essence, our solution offers full explainability while leveraging the efficiencies of contemporary computational techniques. 
	
	{\bf \it Back to Why -- Applications:} Although this manuscript does not focus on specific applications (except for a brief discussion of image generation in Section \ref{sec:GT-sampling}), our motivation is driven by the potential for applied mathematics tools from non-equilibrium statistical mechanics and stochastic optimal control to contribute powerful analytic and computational capabilities to AI. We envision this work opening doors for application-driven researchers—whether physicists developing reduced models for fluid flows, engineers creating synthetic energy systems for computational experimentation, or chemists designing complex reaction pathways—to incorporate their domain-specific expertise into SOC control formulations (e.g., Eq.~(\ref{eq:SOC-A})). In these formulations, \( V(t;{\bm x}) \), \( {\bm f}(t;{\bm x}) \), and \( {\bm A}(t;{\bm x}\) can be fine-tuned to capture the underlying physical, engineering, or chemical principles, thus providing critical application-specific insights and guidance. This approach moves beyond the `black box' nature of traditional AI models, offering interpretable solutions grounded in the laws of nature, engineering constraints, and chemical dynamics.

	\section{This Manuscript Theoretical Contributions}\label{sec:theory}

Let us assume that we have access to a target probability density function, $p_{\text{target}}: \mathbb{R}^d \to [0,1]$, either given explicitly or represented through samples.
	
In this manuscript, our objective is to develop efficient methods for sampling from $p_{\text{target}}(\cdot)$. These methods build upon and extend theoretical foundations from \textit{Stochastic Optimal Control (SOC)}, a classic mathematical engineering discipline which shares conceptual connections with the emerging field of generative AI models, particularly the \textit{Score-Based Diffusion (SBD) models}.
	
Specifically, we focus on a particular subset of SOC known as \textit{Path-Integral Control (PIC)}, as introduced in \cite{kappen_path_2005}. This approach has strong ties to earlier works on nonlinear filtering and control \cite{mitter_non-linear_1981,di_masi_free_1991}, and was later expanded upon in parallel developments \cite{e_todorov_linearly-solvable_2007}. For further background, we refer to related contributions on PIC \cite{theodorou_relative_2012,dvijotham_unifying_2012,kappen_adaptive_2016}.
	
We demonstrate that by introducing an appropriate Path-Integral Control (PIC) framework, we can derive an explicit solution to the underlying stochastic optimal control (SOC) problem. This solution enables the generation of independent and identically distributed (i.i.d.) samples from the target distribution \( p_{\text{target}}(\cdot) \).
	
Our basic SOC formulation is given by the following optimization problem:
	\begin{gather}\label{eq:SOC}
		\min_{{\bm u}(0\to 1)} \mathbb{E}\left[\int_0^1 dt \left(\frac{|{\bm u}(t)|^2}{2} + V(t;{\bm x}(t))\right) \Big| \text{Eqs.~(\ref{eq:SODE}, \ref{eq:p-target})}\right],
	\end{gather}
where the expectation is taken over the controlled stochastic diffusion process, defined by the Stochastic Ordinary Differential Equation (SODE):
	\begin{gather}\label{eq:SODE}
		t \in [0,1]: \quad d{\bm x}(t) = {\bm u}(t) dt + d{\bm W}(t), \quad 
		{\bm x}(0) = {\bm 0}, \quad \forall i,j = 1,\dots,d:\ \mathbb{E}\left[dW_i(t) dW_j(t)\right] = \delta_{ij} dt,
	\end{gather}
where we have initialized the process at \( {\bm x}(0) = {\bm 0} \) without loss of generality. The term \( {\bm W}(t) \) represents the Wiener process with unit covariance. 
	
Additionally, we impose the constraint that the terminal probability distribution at \( t = 1 \) matches the target distribution:
	\begin{gather}\label{eq:p-target}
		p({\bm x}(1)) = p_{\text{target}}({\bm x}(1)).
	\end{gather}
The quadratic term \( |{\bm u}(t)|^2/2 \) in the cost functional of Eq.~(\ref{eq:SOC}) penalizes control effort, thereby encouraging minimal deviation from the natural diffusion dynamics. This ensures that the controlled process generates the target distribution \( p_{\text{target}} \) in an efficient manner, requiring the least effort to modify the underlying stochastic diffusion.

The scalar function $V(t;{\bm x}(t))$ in the integrand of the cost function in Eq.~(\ref{eq:SOC}) imposes a penalty for being at position ${\bm x}(t)$ at time $t$. Since this penalty accumulates over time (within the integral), it reflects our desire to control the state, ${\bm x}(t)$, along the entire path for $t \in [0,1]$. We choose to work with a time-independent positive-definite quadratic scalar function for $V(t;{\bm x})$.
	
We call our basic model -- one stated in Eq.~(\ref{eq:SOC}) however with $V(t;{\bm x})$ which is a time-independent positive-definite quadratic scalar function -- the {\it \bf  Harmonic Path Integral Diffusion (H-PI-Diff)}, because, as we will see below, its solution is linked to that of a linear partial differential equation (PDE) formally similar to the equation for a quantum harmonic oscillator in imaginary time.

Our first theoretical result, presented in Section \ref{sec:th-H}, provides an explicit solution to the H-PI-Diff problem. The essence of this solution is twofold. First, for a fairly general case with an arbitrary (bounded) potential \( V \), we derive the optimal control that drives the forward evolution of the target distribution from an initial \(\delta\)-function. This control is expressed as a functional of the convolution of the target distribution with a space-time dependent function, explicitly given by the ratio of two Green functions. These Green functions correspond to the forward-in-time drift-diffusion of a perturbation initiated at \( t = 0 \) and the backward-in-time drift-diffusion of the perturbation in the adjoint space, initiated at \( t = 1 \). Second, we show that when \( V \) is a quadratic positive-definite function of the state variable, the Green functions -- and consequently, the optimal control/drift -- can be obtained explicitly. In this case, the logarithms of the Green functions take the form of quadratic expressions in the state variables (current and initial), with coefficients that depend on time and are determined as an explicit function of the Hessian of \( V \). This results in an explicit expression for the optimal control: ${\bm u}^{(*)}(t)= a(t){\bm x}(t)-b(t)\hat{\bm x}(t;{\bm x}(t))$, where ${\bm x}(t)$ is the current state at time $t \in [0,1]$, and $\hat{\bm x}(t;{\bm x}(t))$ is the current weighted state, with the weight expressing marginal probability distribution of optimal stochastic process and functions $a(t), b(t)$ and $\hat{\bm x}(t;{\bm x})$ found analytically via convolutions with the target distribution. 
	
Next, in Section \ref{sec:f-A}, we extend the theory to the {\it \bf Forced Harmonic Path Integral Diffusion (FH-PI-Diff)} model. This extension, inspired by \cite{chernyak_stochastic_2013}, generalizes the SOC problem in Eq.~(\ref{eq:SOC}) as follows:
	\begin{align}\label{eq:SOC-A}
		& \min_{{\bm u}(0 \to 1)} \mathbb{E}\left[\int_0^1 dt \left(\frac{|{\bm u}(t)|^2}{2} + V(t; {\bm x}(t)) + {\bm A}^T(t; {\bm x}(t)) \dot{\bm x}(t)\right) \Big| \text{Eqs.~(\ref{eq:SODE-f}, \ref{eq:p-target})}\right], \\
		\label{eq:SODE-f}
		& t \in [0,1]: \quad d{\bm x}(t) = \left({\bm f}(t; {\bm x}) + {\bm u}(t)\right) dt + d{\bm W}(t),
	\end{align}
	where \( {\bm A} \) and \( {\bm f} \) are referred to as the gauge potential and force, respectively. (In the context of prior work on the inclusion of forcing terms, \( {\bm F} \), in SOC problems of the path-integral type, and how this research connects to the Schrödinger Bridge problem, we refer interested readers to \cite{caluya_wasserstein_2022}, with further discussion provided in Section \ref{sec:prior-pubs}.)

	The motivation for this generalization is to remain within the "exactly-solvable" Harmonic Path Integral (H-PI) framework, which allows us to solve Eq.~(\ref{eq:SOC}), while extending it to a more general setting that permits a non-gradient expression for the force in the stochastic dynamics, and potentially for the control as well. (Note that in physics, a non-gradient force is often referred to as a non-conservative force.)

	Our results for the FH-PI-Diff case are fundamentally analogous to those for the H-PI-Diff case discussed earlier. In the case of a bounded potential \( V \) and nonzero \( {\bm f} \) and \( {\bm A} \), we can define two Green functions, allowing the optimal control/drift to be expressed as a convolution of a function involving these Green functions with the target distribution. When the potential \( V \) is quadratic, and both \( {\bm A} \) and \( {\bm f} \) are affine functions of the state variable, and provided that a special matrix constructed from the Hessian of \( V \) and the Jacobians of \( {\bm f} \) and \( {\bm A} \) is positive definite, the optimal control/drift can be expressed as a gradient of a logarithm of a convolution of a Gaussian (explicitly dependent on the Hessian and the Jacobians) with the target distribution.


In Section \ref{sec:f-A}, we explore the three levels of H-PID integrability introduced in the abstract. The \textbf{top level} corresponds to the general FH-PI-Diff model, where the system consists of nonlinear Hamilton-Jacobi-Bellman (HJB) equations for the backward-evolving cost-to-go function in the Stochastic Optimal Control (SOC) framework (from \( t=1 \) to \( t=0 \)), and the linear Fokker-Planck (FP) equations for the forward-evolving probability distribution. Through the Hopf-Cole (HC) transformation, these equations reduce to a system of linear partial differential equations (PDEs). The \textbf{middle level} of integrability arises under a quadratic, positive-definite potential function \( V \) and affine force and gauge terms, which simplifies the Green functions of the HC-transformed HJB and FP equations to Gaussian forms. This Gaussian structure enables optimal control (drift in the stochastic ODE) to be expressed as a convolution between the target distribution and a Gaussian kernel derived from the Green function. At the \textbf{bottom level} of integrability, we consider the case with a uniform quadratic potential \( V \) and zero force and gauge terms, where the time evolution corresponds directly to that of a quantum system with uniform quadratic potential in imaginary time. All experiments in this manuscript demonstrate this bottom-level integrability, showcasing the practical applications of H-PID's structured approach to stochastic sampling.

	\section{From Theory to Algorithms}\label{sec:th-to-alg}

	The theoretical results presented so far are of significant interest due to their practical and algorithmic implications in two key settings:
	\begin{itemize}
		\item \textbf{Sampling based on the Energy Function:} Assume that the target probability distribution $p_{\text{target}}(\cdot)$ is explicitly known, as in \cite{zhang_path_2022}. In this case, we do not require knowledge of the normalization constant (also known as the partition function). Instead, we focus on the \textit{Energy Function}, which is typically defined as the negative logarithm of the target distribution, up to an additive constant. This approach leverages the structure of the energy landscape for efficient sampling. We provide a discussion of this method in Section \ref{sec:SamplingEnFun}.
		
		\item \textbf{Empirical Sampling based on Ground Truth (GT) Data:} When $p_{\text{target}}(\cdot)$ is not explicitly available but is instead approximated through empirical data or Ground Truth (GT) samples, the distribution can be represented as:
		\begin{gather}\label{eq:emp}
			p_{\text{target}}({\bm x}) \approx p_{\text{emp}}({\bm x}) = \frac{1}{S} \sum_{s=1}^S \delta\left({\bm x} - \bm{y}^{(s)}\right),
		\end{gather}
		where $S$ is the number of samples, and $\bm{y}^{(s)}$ denotes the $s$-th sample. This empirical approach is discussed in Section \ref{sec:GT-sampling}.
	\end{itemize}

	\section{Theory: Solution of the Basic Harmonic Path Integral Diffusion Model}\label{sec:th-H}
	
	Let us introduce an auxiliary SOC problem:
	\begin{gather}\label{eq:SOC-aux}
		\min_{{\bm u}(0\to 1)} \mathbb{E}\left[\phi(x(1))+\int_0^1 dt \left(\frac{|{\bm u}(t)|^2}{2}+V(t;{\bm x}(t))\right)\Big| \text{Eq.~(\ref{eq:SODE},\ref{eq:p-target})}\right],
	\end{gather}
	where the functional constraint on the terminal probability distribution (\ref{eq:p-target}) is substituted by the terminal cost function $\phi({\bm x})$. Transformation from Eq.~(\ref{eq:SOC}) to Eq.~(\ref{eq:SOC-aux}) is achieved by solving Eq.~(\ref{eq:SOC-aux}) for the terminal cost function $\phi({\bm x})$ of a general position and then finding such $\phi({\bm x})$ that Eq.~(\ref{eq:p-target}) is satisfied for the respective optimal solution. Formally, once the optimal control ${\bm u}^*(t)$ (expressed as a function of the current $t$) is found, it defines the Fokker-Planck (FP) equation for the optimal probability distribution of ${\bm x}(t)$:
	\begin{gather}\label{eq:p*-aux}
		\partial_t p^*={\bm \nabla} ({\bm u}^* p^*)+\frac{1}{2}\Delta p^*,\quad p^*(0;{\bm x})=\delta({\bm x}),
	\end{gather}
	which must match the target distribution at $t=1$:
	\begin{gather}\label{eq:opt-match}
		p^*(1;{\bm x})=p_{\text{target}}({\bm x}).
	\end{gather}
	
	To solve Eq.~(\ref{eq:SOC-aux}), we employ the standard dynamic programming approach, developed by Bellman (see e.g., \cite{bellman_dynamic_2003}) for more general stochastic optimal control formulations. We introduce the cost-to-go function:
	\begin{gather}\label{eq:J}
		J(t;{\bm x}(t))=\min_{{\bm u}(t\to 1)}\mathbb{E}\left[\phi({\bm x}(1))+\int_t^1 d\tau \left( \frac{|{\bm u}(t)|^2}{2}+V(t;{\bm x}(t))\right)\right],
	\end{gather}
	and derive the Hamilton-Jacobi-Bellman (HJB) equation:
	\begin{gather}\label{eq:HJB}
		-\partial_t J=V+\frac{1}{2}\left(\Delta J-|{\bm \nabla}J|^2\right),\quad J(1;{\bm x}(1))=\phi({\bm x}(1)),
	\end{gather}
	also observing that the optimal control is the negative gradient of $J(t;{\bm x})$, ${\bm u}^*=-{\bm \nabla} J$. 
	
	Although solving the nonlinear PDE (\ref{eq:HJB}) may appear challenging, remarkably, we find -- following e.g., \cite{barber_optimal_2011} -- that due to a special balance between diffusion and nonlinear terms on the right hand side of the HJB Eq.~(\ref{eq:HJB}) (related to the concept of detailed balance in statistical mechanics), the nonlinear PDE (\ref{eq:HJB}) can be transformed into a linear PDE using the Hopf-Cole transformation \cite{hopf_partial_1950,cole_quasi-linear_1951}, $J(t;{\bm x})=- \log \psi(t;{\bm x})$:
	\begin{gather}\label{eq:psi-eq}
		-\partial_t \psi+V\psi=\frac{1}{2}\Delta \psi,\quad \psi(1;{\bm x})=\exp\left(-\phi({\bm x})\right).
	\end{gather}
	Notice that Eq.~(\ref{eq:psi-eq}) is equivalent to the Schr\"{o}dinger equation describing quantum mechanics (in imaginary time) in a $d$-dimensional potential $V(t;{\bm x})$.
	
	The matching conditions (\ref{eq:p*-aux},\ref{eq:opt-match}) between $\phi(\cdot)$ and $p_{\text{target}}(\cdot)$ become:
	\begin{gather}\label{eq:p*}
		\partial_t p^*+{\bm \nabla} (p^* {\bm \nabla} \log \psi)=\frac{1}{2}\Delta p^*,\quad p^*(0;{\bm x})=\delta({\bm x}),\quad  p^*(1;{\bm x})=p_{\text{target}}({\bm x}).
	\end{gather}
	
	\subsection{Green Functions}\label{sec:GF}
	
	The linearity of Eqs.~(\ref{eq:psi-eq}, \ref{eq:p*}) suggests expressing their solutions in terms of Green functions -- backward in time for Eq.~(\ref{eq:psi-eq}) and forward in time for Eq.~(\ref{eq:p*}). These Green functions can be formulated as follows:
	\begin{align}\label{eq:gen-GF-}
		t\in [1\to 0]:\quad & -\partial_t G_- + V G_- = \frac{1}{2}\Delta G_-, \quad G_-(1;{\bm x};{\bm y}) = \delta({\bm x} - {\bm y}),\\ 
		\label{eq:gen-GF+}
		t\in [0\to 1]:\quad & \partial_t G_+ + V G_+ = \frac{1}{2}\Delta G_+, \quad G_+(0;{\bm x};{\bm y}) = \delta({\bm x} - {\bm y}).
	\end{align}
	By leveraging the linearity of Eq.~(\ref{eq:psi-eq}) and combining it with Eq.~(\ref{eq:gen-GF-}), we obtain:
	\begin{gather}\label{eq:gen-conv}
		\psi(t;{\bm x}) = \int d{\bm y} \, \exp\left(-\phi({\bm y})\right) G_-(t;{\bm x};{\bm y}).
	\end{gather}
	Once the solution for $\psi(t;{\bm x})$ is obtained from Eq.~(\ref{eq:gen-conv}), we can immediately verify, using Eq.~(\ref{eq:gen-GF+}), that the appropriate solution to Eq.~(\ref{eq:p*}) is:
	\begin{gather}\label{eq:gen-p*G+}
		p^*(t;{\bm x}) = G_+\left(t; {\bm x}; {\bm 0}\right) \frac{\psi(t; {\bm x})}{\psi(0; {\bm 0})}.
	\end{gather}
	
	This yields the following expression for \( p_{\text{target}} \) in terms of \( \phi \):
	\begin{gather}
		\label{eq:gen-p-target-phi}
		p_{\text{target}}({\bm x}) = p^*(1;{\bm x}) = G_+\left(1; {\bm x}; {\bm 0}\right) \frac{\psi(1; {\bm x})}{\psi(0; {\bm 0})}= \frac{G_+\left(1; {\bm x}; {\bm 0}\right) \exp(-\phi({\bm x}))}{\int d{\bm y} \, G_-(0; {\bm 0}; {\bm y}) \exp(-\phi({\bm y}))}.
	\end{gather}
	
	Eq.~(\ref{eq:gen-p-target-phi}) can also be inverted to express \( \phi \) (and subsequently \( {\bm u}^* \), using Eq.~(\ref{eq:gen-conv})) in terms of \( p_{\text{target}} \) as follows:
	\begin{align}
		\label{eq:gen-phi-via-target}
		\phi({\bm x}) &= \log\left(\frac{G_+\left(1; {\bm x}; {\bm 0}\right)}{p_{\text{target}}({\bm x})}\right) + \text{const}, \\ \label{eq:gen-u*-via-target}
		u^*(t; {\bm x}) &=  
		\nabla_{\bm x} \log\left(\int d{\bm y} \, \exp\left(-\phi({\bm y})\right) G_-(t; {\bm x}; {\bm y})\right) = \nabla_{\bm x} \log\left(\int d{\bm y} \, p_{\text{target}}({\bm y}) \frac{G_-(t; {\bm x}; {\bm y})}{G_+\left(1; {\bm y}; {\bm 0}\right)}\right).
	\end{align}
	
	The above formulas are general and apply to arbitrary (and potentially time-dependent) potentials $V(t; {\bm x})$. However, solving the effective "quantum mechanics" governing Eqs.~(\ref{eq:gen-GF-}, \ref{eq:gen-GF+}) analytically is typically impossible when the potential $V$ is of general form.
	
	Thus, we now turn our attention to identifying integrable cases where the corresponding quantum mechanics equations can be solved analytically. Naturally, the first case we will explore is potential-free quantum mechanics, i.e., the case where $V = 0$.
	
	\subsection{Zero Potential, $V=0$} \label{sec:V=0}
	
	In the potential-free case, the Green functions introduced earlier in Section \ref{sec:GF} reduce to simple diffusion kernels, which are Gaussians. Consequently, we arrive at the following simplified forms of the general equations:
	\begin{align}\label{eq:Rag1}
		\forall t \in [0,1]\quad &\psi(t;{\bm x}) = \int d{\bm y} \, {\cal N}\left({\bm y}|{\bm x};(1-t){\bm 1}\right)\exp\left(-\phi({\bm y})\right),\\
		\label{eq:Rag2}
		\forall t,\tau \in [0,1], \ \tau < t\quad & p^*(t;{\bm x}|\tau;{\bm y}) = {\cal N}\left({\bm x}|{\bm y};(t-\tau){\bm 1}\right)\frac{\psi(t;{\bm x})}{\psi(\tau;{\bm y})},\\
		\label{eq:Rag3}
		& p_{\text{target}}({\bm x}) = \frac{{\cal N}\left({\bm x}|{\bm 0};{\bm 1}\right)\exp(-\phi({\bm x}))}{ \int d{\bm y} \, {\cal N}\left({\bm y}|{\bm 0};{\bm 1}\right) \exp(-\phi({\bm y}))},\\
		\label{eq:Rag4}
		& \phi({\bm x}) = \log\left(\frac{{\cal N}\left({\bm x}|{\bm 0};{\bm 1}\right)}{p_{\text{target}}({\bm x})}\right) + \text{const},\\
		\label{eq:Rag5}
		& u^*(t;{\bm x}) = 
		\nabla_{\bm x} \log\left(\int d{\bm y} \, {\cal N}\left({\bm y}|{\bm x};{\bm 1}(1-t)\right)\frac{p_{\text{target}}({\bm y})}{{\cal N}\left({\bm y}|{\bm 0};{\bm 1}\right)}\right).
	\end{align}
	
	These results establish an explicit relationship between \( p_{\text{target}}(\cdot) \) and \( \phi(\cdot) \) in the $V=0$ case. This relationship was derived earlier in Theorem 2.1 of \cite{tzen_theoretical_2019}, with additional references to \cite{pavon_stochastic_1989, dai_pra_stochastic_1991}.
	
	Eq.~(\ref{eq:Rag5}) serves as a foundational starting point for developing novel algorithmic approaches to sampling from \( p_{\text{target}}(\cdot) \), which will be further explored in Section \ref{sec:SamplingEnFun}.

	\subsection{Quadratic Potential}
	
	Next, we consider the case where $V \neq 0$, focusing specifically on a potential $V({\bm x})$ that is a quadratic, positive-definite scalar function of ${\bm x}$. 
 
	\subsubsection{General Quadratic Potential}
	
	Consider the case of a time-independent quadratic potential in general form, \( V({\bm x}) = \frac{1}{2} {\bm x}^T \hat{\bm \beta} {\bm x} \). In this scenario, as detailed in Appendices \ref{sec:GF-gen} and \ref{sec:GF+gen}, the Green functions for the reversed and forward processes can be obtained analytically. The key insight here is that the Green functions, \( G_{\pm}({\bm x}; {\bm y}; t) \), are Gaussian. Moreover, the logarithms of these Green functions are quadratic forms, with time-dependent coefficients provided explicitly in the appendices. Once these explicit expressions for the Green functions are derived, substituting them into the corresponding equations in Section \ref{sec:GF} allows us to express the optimal control field as an explicit functional of the target distribution.
	
	To save space, we present here the final expression for the ratio of the Green functions, which enters the expression for the optimal control field (according to Eq.~(\ref{eq:gen-u*-via-target})) for the case of a uniform quadratic potential, \( V({\bm x}) = \frac{\beta}{2} |{\bm x}|^2 \). The detailed derivations of the Green functions for this case are provided in Appendices \ref{sec:GF-} and \ref{sec:GF+}.
	\begin{gather}\label{eq:G-ratio-uni}
		\frac{G_-(t; {\bm x}; {\bm y})}{G_+\left(1; {\bm y}; {\bm 0}\right)} =
	  \sqrt{\frac{\sinh\left(\sqrt{\beta}\right)}{\sinh\left((1-t)\sqrt{\beta}\right)}}
	\exp\left(-\frac{\sqrt{\beta}}{2}
	\left(({\bm x}^2+{\bm y}^2)\coth\left((1-t)\sqrt{\beta}\right)-{\bm y}^2\coth\left(\sqrt{\beta}\right)-\frac{2({\bm x}^T{\bm y})}{\sinh\left((1-t)\sqrt{\beta}\right)}\right)
	\right)
	\end{gather}
	
	\section{Theory: Forced Harmonic Path Integral Diffusion}\label{sec:f-A}
	
	This section is devoted to solving Eqs.~(\ref{eq:SOC-A}). We follow a similar approach as before, by formulating and solving the corresponding HJB equation, which generalizes Eq.~(\ref{eq:HJB}):
	\begin{gather}\label{eq:HJB-A-f}
		-\partial_t J = V + \frac{1}{2} \left({\bm \nabla}^T({\bm \nabla} J + {\bm A}) - |{\bm \nabla} J + {\bm A}|^2\right) + {\bm f}^T({\bm \nabla} J + {\bm A}),
	\end{gather}    
	and the optimal control becomes:
	\begin{gather}\label{eq:opt-control-J-A}
		{\bm u}^* = -{\bm \nabla} J - {\bm A}.
	\end{gather}
	Notice that, in general, when the gauge potential ${\bm A} \neq 0$, the control field is not a gradient of a scalar function.
	
	Substituting \( J(t; {\bm x}) = - \log \psi(t; {\bm x}) \) into Eq.~(\ref{eq:HJB-A-f}), as before, we arrive at:
	\begin{align}\label{eq:psi-eq-A-f}
		& -\partial_t \psi + \tilde{V} \psi + \tilde{\bm A}^T {\bm \nabla} \psi = \frac{1}{2} \Delta \psi,\quad \psi(1; {\bm x}) = \exp\left(-\phi({\bm x})\right), \\
		\label{eq:tilde}
		& \tilde{V} \doteq V + \frac{1}{2} {\bm \nabla}^T {\bm A} + {\bm f}^T {\bm A} - \frac{1}{2} |{\bm A}|^2, \quad \tilde{\bm A} \doteq {\bm A} - {\bm f}.
	\end{align}
	Then, the Fokker-Planck equation for the optimal control becomes:
	\begin{gather}\label{eq:p*-A-f}
		\partial_t p^* + {\bm \nabla}^T\left( p^* ({\bm \nabla} \log \psi - \tilde{\bm A})\right) = \frac{1}{2} \Delta p^*, \quad p^*(0; {\bm x}) = \delta({\bm x}), \quad p^*(1; {\bm x}) = p_{\text{target}}({\bm x}).
	\end{gather}
	
	Following the logic detailed in Section \ref{sec:GF} for the case where \( {\bm A} = {\bm f} = 0 \), we introduce the Green functions for the general case, both reversed in time and forward in time. These Green functions satisfy the following equations:
	\begin{align}\label{eq:FA-GF-}
		t \in [1 \to 0]: \quad & -\partial_t G_- + \tilde{V}({\bm x}; t) G_- + \tilde{\bm A}^T {\bm \nabla} G_- = \frac{1}{2} \Delta G_-, \quad G_-(1; {\bm x}) = \delta({\bm x} - {\bm y}), \\
		\label{eq:FA-GF+}
		t \in [0 \to 1]: \quad & \partial_t G_+ + \tilde{V}({\bm x}; t) G_+ - {\bm \nabla}^T (\tilde{\bm A} G_+) = \frac{1}{2} \Delta G_+, \quad G_+(0; {\bm x}) = \delta({\bm x} - {\bm y}).
	\end{align}
	
It is straightforward to verify that all the equations derived in Section \ref{sec:GF}, except Eqs.~(\ref{eq:gen-GF-}, \ref{eq:gen-GF+}) for the Green functions, apply to the case of an arbitrary and non-zero \( {\bm A} \) and \( {\bm f} \), with Eqs.~(\ref{eq:FA-GF-}, \ref{eq:FA-GF+}) replacing the Green function equations.
	
Based on our previous experience with the case \( {\bm A} = {\bm f} = 0 \) we discover that, if we choose \( V \) as a quadratic form in \( {\bm x} \), and consider \( {\bm A} \) and \( {\bm f} \) as affine vector functions of \( {\bm x} \), while requiring that the resulting quadratic \( \tilde{V} \) is positive definite, we observe that the {\bf Gaussian ansatz for the Green functions holds}. This allows us to extend the integrable construction, based on Gaussian Green functions, to this more general case as well.

The Gaussian nature of the Green's functions implies that $\log\left(\frac{G_-(t;{\bm x};{\bm y})}{G_+(1;{\bm y};{\bm 0})}\right)$ takes the form of a quadratic expression in both ${\bm x}$ and ${\bm y}$. This allows us to make a very general statement about the structure of the optimal control. Specifically, we derive:
\begin{gather}\label{eq:u*-Gauss}
    u^*(t;{\bm x}(t)) = \nabla_{\bm x} \log\left(\int d{\bm y}\ p_{\text{target}}({\bm y})
    \frac{G_-(t;{\bm x}(t);{\bm y})}{G_+(1;{\bm y};{\bm 0})}\right) = a(t){\bm x}(t) - b(t)\hat{\bm x}(t;{\bm x}(t)),
\end{gather}
where $a(t)$, $b(t)$, and $\hat{\bm x}(t; {\bm x})$ are explicit functionals of $p_{\text{target}}(\cdot)$, and of the only time dependent expressions induced by \( V \), \( {\bm A} \) and \( {\bm f} \). Determining $a(t)$, $b(t)$, and $\hat{\bm x}(t;{\bm x})$ in the general case requires solving a set of ordinary differential equations (in time only), the analysis of which we leave for future studies.

\section{Experiments: Sampling based on the Energy Function}\label{sec:SamplingEnFun}
	
The explicit solution of the H-PI Diffusion problem, given by Eqs.~(\ref{eq:gen-phi-via-target}, \ref{eq:gen-u*-via-target}), can be further utilized for sampling purposes -- this is the main topic of this section.
	
Consider the case where $p_{\text{target}}(\cdot)$ is known up to a normalization factor:
	\begin{gather}\label{eq:energy}
		p_{\text{target}}({\bm x})=\frac{\exp(-E({\bm x}))}{Z},\quad Z=\int d{\bm y} \exp(-E({\bm y})),
	\end{gather}
where $E(\cdot)$ is the so-called energy function, and $Z$ is the generally unknown and computationally challenging normalization constant, often referred to as the partition function. Substituting this into Eq.~(\ref{eq:gen-u*-via-target}) yields:
	\begin{gather}\label{eq:u*-energy}
		u^*(t;{\bm x}) = \nabla_{\bm x} \log\left(\int d{\bm y}\ \exp(-E({\bm y}))
		\frac{G_-(t;{\bm x};{\bm y})}{G_+(1;{\bm y};{\bm 0})}\right).
	\end{gather}
	
We aim to leverage the explicit formula in Eq.~(\ref{eq:u*-energy}) to run the stochastic dynamics in Eq.~(\ref{eq:SODE}), replacing ${\bm u}(\cdot,\cdot)$ with ${\bm u}^*(\cdot,\cdot)$, thereby generating i.i.d. samples from $p_{\text{target}}(\cdot)$ at $t=1$. However, this requires a computationally efficient method to evaluate $u^*(t;{\bm x})$ multiple times, which is the focus of the subsequent discussion.
	
Fortunately, the unknown normalization constant $Z$ does not appear directly in Eq.~(\ref{eq:u*-energy}). The main challenge, however, lies in the evaluation of the integral over ${\bm y}$. In the following two subsections, we develop two approaches to overcome this difficulty using {\it \bf Importance Sampling} (IS): one universal method that is independent of the specific energy function, and another approach that explicitly depends on the form of the energy function.

	\subsection{Energy Function Independent (Universal) Importance Sampling} \label{sec:IS-universal}
	
When $t$ is sufficiently close to $1$, the dependence of the integrand on ${\bm y}$ is dominated by the extremum of its "universal" (i.e., energy-function-independent) part. In this case, we can rely on the following approximation:
	\begin{gather}\label{eq:univ}
		\int d{\bm y }\  \exp(-E({\bm y}))
		\frac{G_-(t;{\bm x};{\bm y})}{G_+(1;{\bm y};{\bm 0})}\approx \frac{G_-(t;{\bm x};{\bm y}^*)}{G_+(1;{\bm y}^*;{\bm 0})}\frac{\exp(-E({\bm y}^*))}{\sqrt{(2\pi)^d \det(\hat{\bm H})}},
	\end{gather}
where the $t$- and ${\bm x}$-dependent quantities ${\bm y}^*$ and $\hat{\bm H}$ are defined by:
	\begin{align} \label{eq:y*}
		\forall t,{\bm x}: &\quad \nabla_{\bm y}\log\left(\frac{G_-(t;{\bm x};{\bm y})}{G_+(1;{\bm y};{\bm 0})}\right)\Bigg|_{{\bm y}\to {\bm y}^*}=0,\\ \label{eq:H}
		& i,j=1,\cdots,d:\quad H_{ij}(t;{\bm x})=-\partial_{y_i}\partial_{y_j}\log\left(\frac{G_-(t;{\bm x};{\bm y})}{G_+(1;{\bm y};{\bm 0})}\right)\Bigg|_{{\bm y}\to {\bm y}^*}.
	\end{align}
	
A notable feature of the "Gaussian Integrability" explored in this manuscript is that the logarithms of the Green functions form a joint quadratic expression in ${\bm x}$ and ${\bm y}$, with coefficients dependent on $t$. Consequently, ${\bm y}^*(t;{\bm x})$ is linear in ${\bm x}$, and $\hat{\bm H}(t;{\bm x})$ remains time-independent. 
	
While Eq.~(\ref{eq:univ}) is universal (energy-function-independent) and straightforward to compute, it remains an approximation -- a stationary-point approximation, to be exact. We expect this approximation to hold asymptotically as $t\to 1$, where the integrand in Eq.~(\ref{eq:univ}) approaches a $\delta$-function and the Hessian $\hat{\bm H}$ becomes singular, i.e., $\hat{\bm H}\to \infty$. However, since we have an explicit expression for the Hessian at all $t<1$ and since we expect that the Gaussian approximation for the integrand (as a function of ${\bm y})$ will be quite reasonable at smaller values of $t$, we propose constructing a universal IS scheme using a Gaussian probe function centered at ${\bm y}^*$, with covariance matrix $\hat{\bm H}^{-1}$ (precision matrix $\hat{\bm H}$). Specifically, we derive:
	\begin{gather}\label{eq:IS}
		\int d{\bm y }\  \exp(-E({\bm y}))
		\frac{G_-(t;{\bm x};{\bm y})}{G_+(1;{\bm y};{\bm 0})}=
		\mathbb{E}_{{\bm y}\sim {\cal N}\left(\cdot;{\bm y}^*;\hat{\bm H}^{-1}\right)}\left[\frac{\exp(-E({\bm y}))}{{\cal N}\left({\bm y};{\bm y}^*;\hat{\bm H}^{-1}\right)}
		\frac{G_-(t;{\bm x};{\bm y})}{G_+(1;{\bm y};{\bm 0})}\right].
	\end{gather}
	
Let us now adopt the general formulas to the special case of the zero force and vector field, ${\bm F}={\bm A}=0$ and isotropic quadratic potential, $V(t;x)=\beta |{\bm x}|^2/2$.

\subsubsection{Zero Field, Isotropic Quadratic Potential}
	
Explicit expressions for the ratio of the Green functions, ${\bm y}^*(t;{\bm x})$ and $\hat{\bm H}(t;{\bm x})$ for the case ${\bm A}={\bm F}=0$ and $V(t;x)=\beta |{\bm x}|^2/2$ are provided in Appendix \ref{app:uni-IS} -- see Eqs.~(\ref{eq:G-ratio},\ref{eq:y*-2}) and (\ref{eq:H-1}), respectively. Substituting these expressions in Eqs.~(\ref{eq:IS},\ref{eq:u*-energy}), we arrive at the following universal IS formula for the optimal control:
	\begin{align}\nonumber 
		u^*(t;{\bm x}) & =\nabla_{\bm x}\log\left(\int d{\bm y }\  \exp(-E({\bm y}))
		\frac{G_-(t;{\bm x};{\bm y})}{G_+(1;{\bm y};{\bm 0})}\right)=\frac{\int d{\bm y }\  \exp(-E({\bm y}))\frac{\nabla_{\bm x} G_-(t;{\bm x};{\bm y})}{G_+(1;{\bm y};{\bm 0})}}{\int d{\bm y }'\  \exp(-E({\bm y}'))\frac{G_-(t;{\bm x};{\bm y}')}{G_+(1;{\bm y'};{\bm 0})}}\\ \label{eq:IS-u} & =\frac{\sqrt{\beta}}{\sinh((1-t)\sqrt{\beta})}
		\frac{\int d{\bm y }\  \left({\bm y}-{\bm x}\cosh((1-t)\sqrt{\beta})\right)\exp(-E({\bm y}))\frac{G_-(t;{\bm x};{\bm y})}{G_+(1;{\bm y};{\bm 0})}}{\int d{\bm y }'\  \exp(-E({\bm y}'))\frac{G_-(t;{\bm x};{\bm y}')}{G_+(1;{\bm y}';{\bm 0})}}\\ 
		& =
		\frac{\sqrt{\beta}}{\sinh((1-t)\sqrt{\beta})}\frac{
			\mathbb{E}_{{\bm y}\sim {\cal N}\left(\cdot;{\bm y}^*;\hat{\bm H}^{-1}\right)}\left[\left({\bm y}-{\bm x}\cosh((1-t)\sqrt{\beta})\right)\frac{\exp(-E({\bm y}))}{{\cal N}\left({\bm y};{\bm y}^*;\hat{\bm H}^{-1}\right)}
			\frac{G_-(t;{\bm x};{\bm y})}{G_+(1;{\bm y};{\bm 0})}\right]}{\mathbb{E}_{{\bm y}'\sim {\cal N}\left(\cdot;{\bm y}^*;\hat{\bm H}^{-1}\right)}\left[\frac{\exp(-E({\bm y}'))}{{\cal N}\left({\bm y}';{\bm y}^*;\hat{\bm H}^{-1}\right)}
			\frac{G_-(t;{\bm x};{\bm y}')}{G_+(1;{\bm y}';{\bm 0})}\right]} \nonumber \\  & \approx  
		\frac{\sqrt{\beta}}{\sinh((1-t)\sqrt{\beta})}\frac{
			\sum_{n=1}^N\left[\left({\bm y}^{(n)}-{\bm x}\cosh((1-t)\sqrt{\beta})\right)\frac{\exp(-E({\bm y}^{(n)}))}{{\cal N}\left({\bm y}^{(n)};{\bm y}^*;\hat{\bm H}^{-1}\right)}
			\frac{G_-(t;{\bm x};{\bm y}^{(n)})}{G_+(1;{\bm y}^{(n)};{\bm 0})}\right]}{\sum_{m=1}^N\left[\frac{\exp(-E({\bm y}^{(m)}))}{{\cal N}\left({\bm y}^{(m)};{\bm y}^*;\hat{\bm H}^{-1}\right)}
			\frac{G_-(t;{\bm x};{\bm y}^{(m)})}{G_+(1;{\bm y}^{(m)};{\bm 0})}\right]}, \label{eq:UHIS}
	\end{align}
where $y^{(n)}$, with $n=1,\cdots,N$, are $N$ i.i.d. samples generated from ${\cal N}\left(\cdot;{\bm y}^*;\hat{\bm H}^{-1}\right)$. Note that according to Eq.~(\ref{eq:H-1}) the Hessian matrix is proportional to the identity matrix, and therefore its inverse is also proportional to the identity matrix and straightforward to compute.
	
We expect that only $N=O(1)$ samples within the universal IS framework are required to achieve sufficient accuracy for all $t$ and ${\bm x}$. The universal nature of this IS scheme allows pre-computation of samples, which can then be reused for different energy functions—an invaluable feature for applications that require sampling from target distributions with varying energy functions or landscapes.
	
\subsection{Non-Universal Stationary Point Approximation for Optimal Control}
	\label{sec:IS-nonuniversal}
	
	\begin{figure}
		\begin{algorithm}[H] 
			\caption{UHIS algorithm to generate i.i.d. samples from the target distribution and estimate $Z$.}
			\label{alg:Z}
			\begin{algorithmic}[1]
				\Require Number of samples $S$; temporal discretization of Eq.~(\ref{eq:SODE}); $\beta$; energy function $E({\bm x})$; number of samples $N$ in Eq.~(\ref{eq:UHIS})
				\State $s = 0$
				\While{$s < S$}
				\State $s = s + 1$
				\State Solve Eq.~(\ref{eq:SODE}) with ${\bm u}(t)$ substituted by ${\bm u}^{(*)}(t; {\bm x}(t))$ from Eq.~(\ref{eq:UHIS})
				\State $\bm{y}^{(s)} = {\bm x}(1)$
				\EndWhile
				\State $Z = \frac{1}{S}\sum_{s=1}^{S} \exp(-E(\bm{y}^{(s)}))$ 
			\end{algorithmic}
		\end{algorithm}
	\end{figure}
	
The stationary point approximation, and the IS discussed in Section \ref{sec:IS-universal}, can also be improved by accounting for the energy function factor in the integrand on the left hand side of Eq.~(\ref{eq:IS}). In this case the non-universal (as now dependent on the energy function, $E({\bm x})$) stationary-point versions of Eqs.~(\ref{eq:y*},\ref{eq:H}) become
	\begin{align} \label{eq:y-E}
		\forall t,{\bm x}: &\quad \nabla_{\bm y}\left(\log\left(\frac{G_-(t;{\bm x};{\bm y})}{G_+(1;{\bm y};{\bm 0})}\right)- E({\bm y})\right)\Bigg|_{{\bm y}\to {\bm y}^{\diamond}}=0,\\ \label{eq:H-E}
		& i,j=1,\cdots,d:\quad H_{ij}(t;{\bm x})=    \partial_{y_i}\partial_{y_j}\left(\log\left(\frac{G_-(t;{\bm x};{\bm y})}{G_+(1;{\bm y};{\bm 0})}\right)-E({\bm y})\right)\Bigg|_{{\bm y}\to {\bm y}^{\diamond}}.
	\end{align}
	
Computation of the stationary-point in this non-universal case is more complicated (equivalent to solving an optimization problem), however we may conjecture  that in this case we can get away only with a stationary point estimate (only stationary point,  and not IS around it --- even though we can later build IS around the non-universal stationary-point too):
	\begin{align}\nonumber
		u^*(t;{\bm x}) & \approx \sinh ((1-t)\sqrt{\beta})\sqrt{\beta} \ \nabla_{\bm x} \sup_{\bm y} \left(({\bm x}^T{\bm y}) - \frac{E({\bm y})+|{\bm y}|^2 \left(\coth \left((1-t)\sqrt{\beta}\right)-\coth (\sqrt{\beta})\right)\sqrt{\beta}/2}{\sinh ((1-t)\sqrt{\beta})\sqrt{\beta}}\right)\\
		\label{eq:inst}  & = \sinh ((1-t)\sqrt{\beta})\sqrt{\beta}\ \text{arg}\sup_{\bm y} \left(({\bm x}^T{\bm y}) - \frac{E({\bm y})+|{\bm y}|^2 \left(\coth \left((1-t)\sqrt{\beta}\right)-\coth (\sqrt{\beta})\right)\sqrt{\beta}/2}{\sinh ((1-t)\sqrt{\beta})\sqrt{\beta}}\right),
	\end{align}
	therefore requiring computing (re-scaled) Legendre-Fenchel transform of the energy function.
	
	\subsection{Gaussian Mixture Test}\label{eq:GaussMixt}

	\begin{figure}
		\centering
		\begin{adjustbox}{valign=c,clip,trim=0.75in 6in 0in 0in} 
			\includegraphics[scale=1]{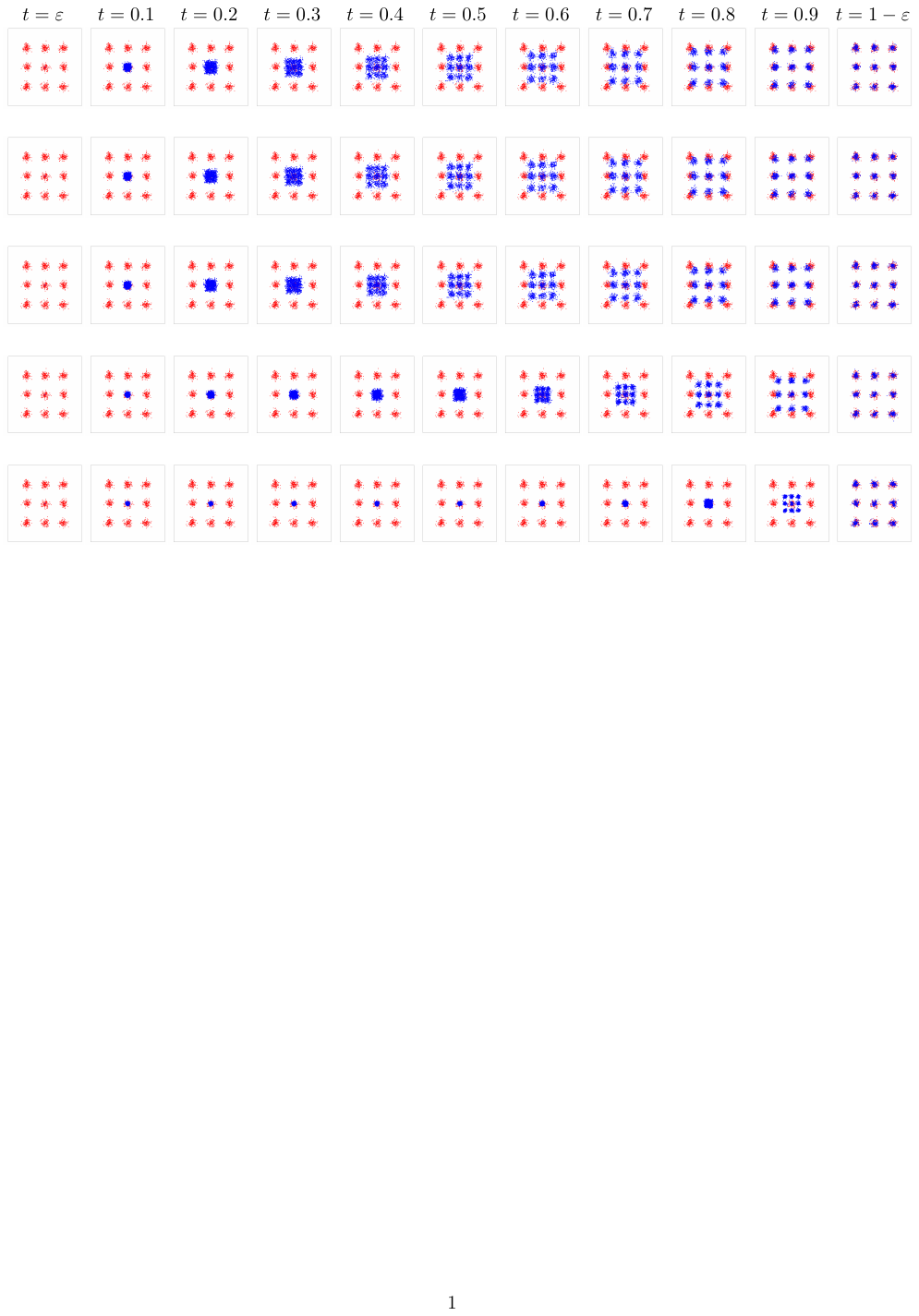}
		\end{adjustbox}
		\caption{ The red circles represent i.i.d. samples drawn from the target distribution, which is a mixture of nine Gaussian components arranged in a \((3 \times 3)\) square grid, where distance between nearest grid points is $5$, variance of the Gaussians is 0.5. The blue circles illustrate the temporal evolution of samples governed by Eq.~(\ref{eq:SODE}), where the time interval \( t = (0, 1) \) is discretized into 200 steps and the optimal control \({\bm u}(t; {\bm x}) \to {\bm u}^*(t; {\bm x})\) is defined according to Eq.~(\ref{eq:UHIS}) with \(N=10000\). Each row corresponds to a different value of $\beta$ ($\beta = 0$, $\beta = 0.1$, $\beta = 1$, $\beta = 10$, and $\beta = 100$, respectively).
  }
		\label{fig:cl2}
	\end{figure}

	\begin{figure}[h!]
		\subfigure[]{
			\centering
			\includegraphics[scale=0.4]{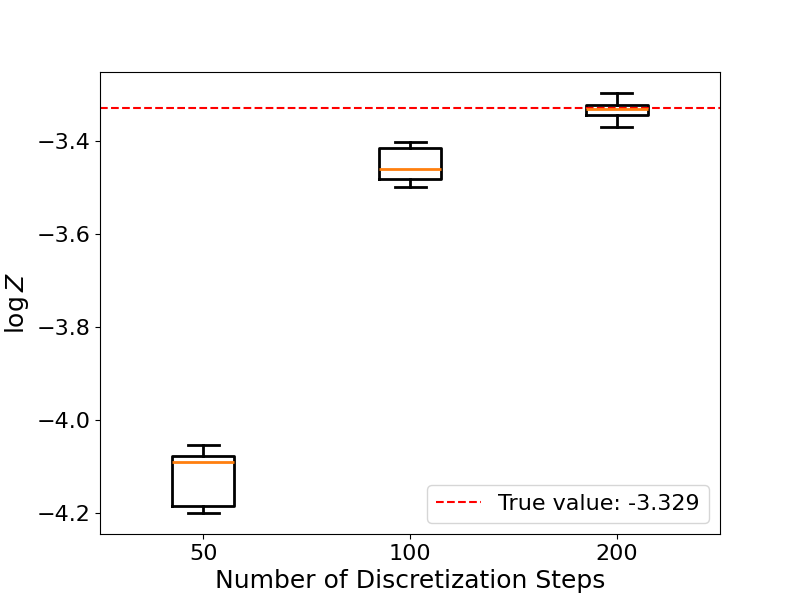} }
		\subfigure[]{
			\centering
			\includegraphics[scale=0.4]{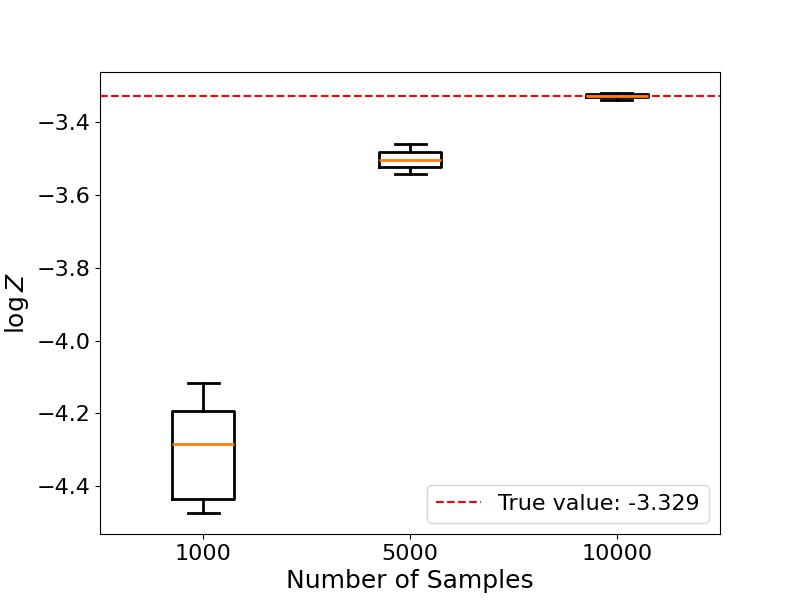} }
		\vspace{-5pt}
		\caption{
			Estimation of the partition function, $Z$, as a function of (a) the number of time-discretization steps and (b) the number of samples, shown in Subfigures (a) and (b) respectively. The partition function $Z$ is calculated according to Algorithm (\ref{alg:Z}) in a setup similar to Figs.~\ref{fig:cl2}, where the target distribution is a Gaussian mixture. The distance between adjacent centers in the $3 \times 3$ Gaussian mixture grid is 5, with $\beta = 0.5$. Both Subfigures present the mean and variance as boxplots, based on 10 independent experiments.
		}
		\label{fig:boxplot} 
	\end{figure}
	
Eq.~(\ref{eq:SOC}) suggests a practical IS implementation of the the Harmonic Path Integral Diffusion in its isotropic version associated with Eq.~(\ref{eq:UHIS}). We coin the algorithm, shown in Algorithm (\ref{alg:Z}) Universal Harmonic Importance Sampling (UHIS),  therefore emphasizing its ``universality" -- that is the fact that the algorithm does not depend on the explicit form of the target energy function, ${\bm E}({\bm x})$.

In the remainder of this Subsection, we evaluate the performance of the UHIS Algorithm (\ref{alg:Z}) using a Gaussian mixture model on a two-dimensional grid -- a standard synthetic benchmark commonly employed in state-of-the-art studies on diffusion model training for sampling from distributions defined via an energy function (e.g., \cite{richter_improved_2024}). This test serves a dual purpose: first, it presents a challenging task for general-purpose algorithms due to the rugged energy landscape; second, it has a simple specialized solution if one possesses additional knowledge of the energy landscape's minima location and intensity. Notably, this specialized knowledge is not utilized in the UHIS algorithm being tested. Instead, it serves as a reference to assess the quality and effectiveness of UHIS in navigating complex energy landscapes without any prior knowledge.

Performance of UHIC Algorithm (\ref{alg:Z}) is illustrated in Fig.~\ref{fig:cl2}, where blue dots show samples produced by the specialized algorithm and the red dots show samples from the target distribution as a reference at different times. In all the tests we track evolution of $1000$ (blue) samples of UHIC in 200 equally spaced time intervals (discretizing time evolution which is continuous in theory). Different rows of Fig.~\ref{fig:cl2} show UHIC algorithm evolution for different values of the parameter $\beta$ changing the strength of the quadratic potential (relative to the control efforts term, $|{\bm u}|^2$, in the SOC objective).

The empirical estimation of the partition function $Z$, concluding Algorithm \ref{alg:Z}, is quantitatively assessed by comparing it to the exact value obtained using a specialized algorithm, as illustrated in Fig.~\ref{fig:boxplot}. This comparison provides a robust quantitative measure of the quality assurance for our primary objective -- generating i.i.d. samples from the target distribution.

When all parameters are properly selected and UHIS successfully achieves its goal of sampling from the target distribution, the optimal stochastic dynamics can generally be divided into two distinct phases. In the initial phase, the samples spread from zero, although not necessarily widely across the domain. By the end of this phase -- which, as shown in Fig.~\ref{fig:cl2}, lasts for a duration that depends on $\beta$ -- a \textbf{structure} reminiscent of the target distribution \textbf{begins to emerge}. This is followed by the second stage, where the structure grows, eventually transforming into the target distribution.

\begin{figure}[h!]
		\subfigure[]{
			\centering
			\includegraphics[scale=0.25]{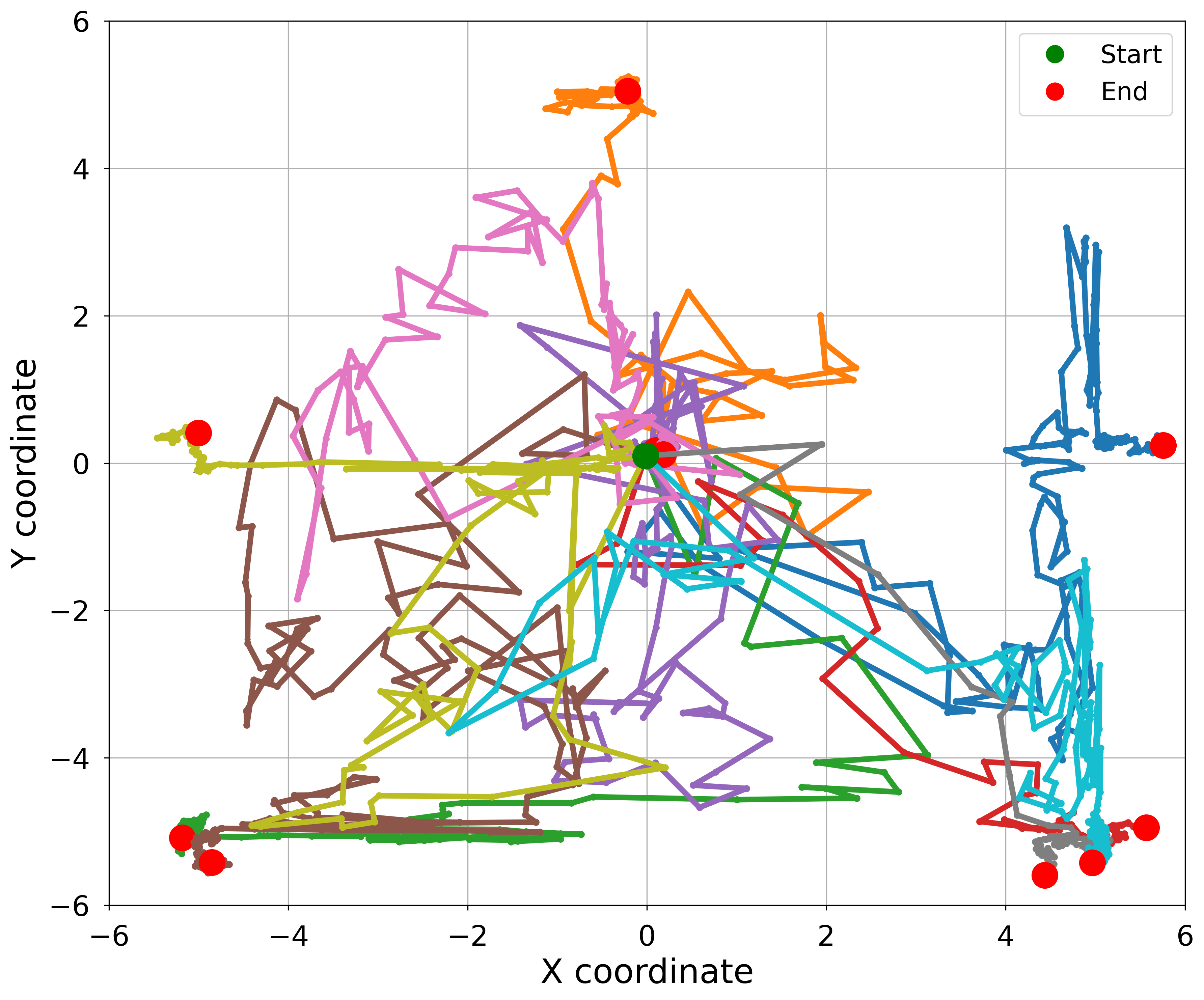} }
		\subfigure[]{
			\centering
			\includegraphics[scale=0.25]{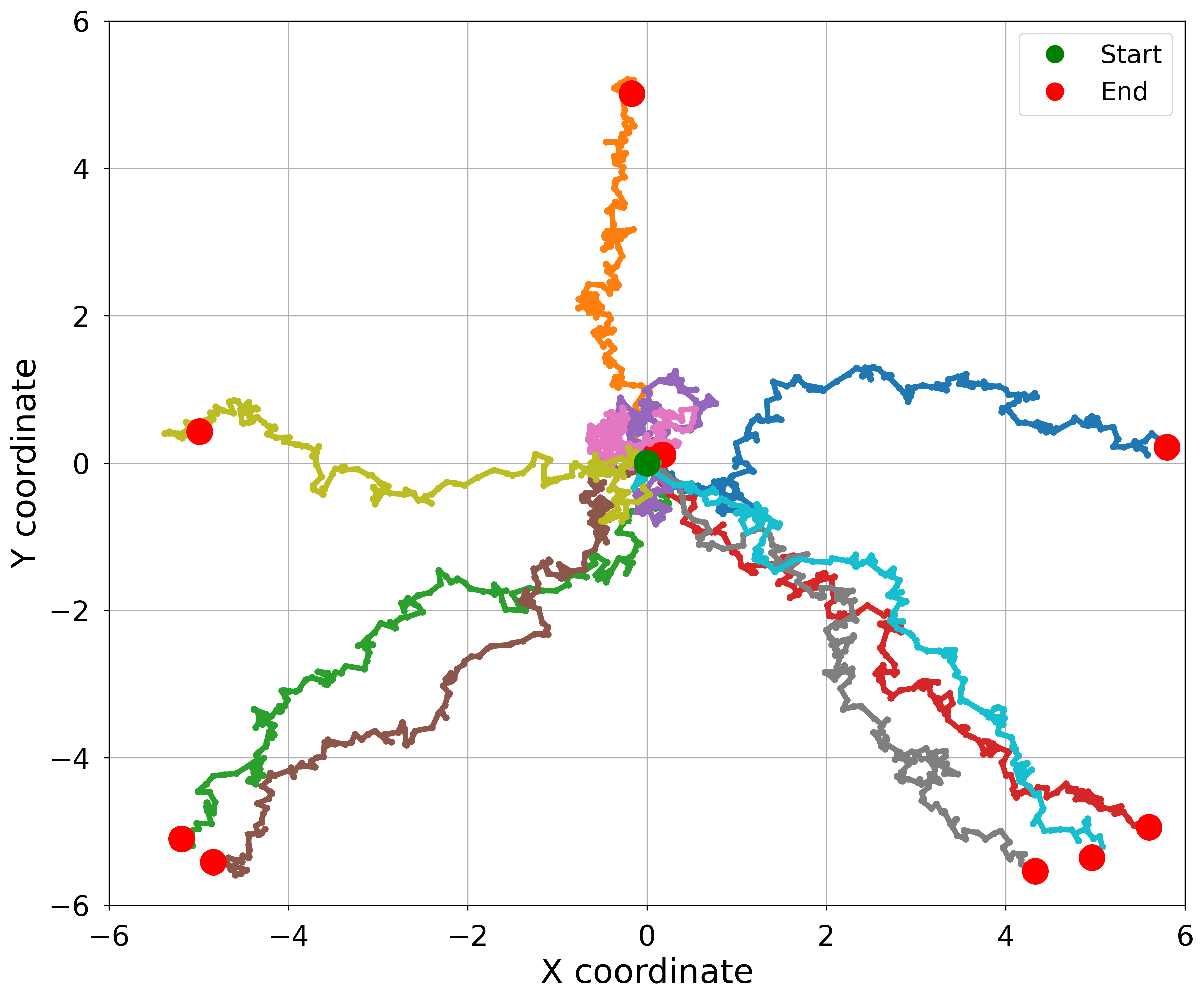} }
		\vspace{-5pt}
		\caption{Trajectories of 10 samples generated according to the optimal stochastic dynamics for the target distribution corresponding to the main use case of the $3 \times 3$ grid. The samples start at the origin and evolve over time in the weighted state space (a-left) $\hat{\bm x}(t; {\bm x}(t))$ and the state space (b-right) $\bm{x}(t)$. Notice that exploration in the weighted state space (left) is more extensive, with a greater variety of states visited during the process. Here $\beta=0.1$ and number discretization steps is $200$. } 
		\label{fig:particles_trajectory} 
\end{figure}

A more detailed analysis of the dynamic phase transition reveals several key observations:
\begin{enumerate}
\item Interpreting the nine Gaussians in the mixture as analogous to different species, we can view the dynamic transition as a speciation-like process, similar to the dynamic speciation transition discussed in \cite{biroli_dynamical_2024}. In this context, a single species (corresponding to one of the nine centers) is clearly selected from the mixture. 

\item The structure of Eq.~(\ref{eq:IS-u}) suggests studying what we coin the {\bf current weighted state}:
\begin{gather}\label{eq:hat-x}
    \hat{{\bm x}}(t;{\bm x}(t)) \doteq \frac{\int d{\bm y }\ {\bm y}\exp(-E({\bm y}))\frac{G_-(t;{\bm x}(t);{\bm y})}{G_+(1;{\bm y};{\bm 0})}}{\int d{\bm y }'\  \exp(-E({\bm y}'))\frac{G_-(t;{\bm x}(t);{\bm y}')}{G_+(1;{\bm y}';{\bm 0})}},
\end{gather}
and specifically, how it evolves over time from $t = 0$ to $t = 1$, eventually producing a sample from the target distribution at $t = 1$. This analysis is shown in Fig.~\ref{fig:cl3} of Appendix \ref{app:figs}, which extends Fig.~\ref{fig:cl2} by adding the evolution of multiple $\hat{{\bm x}}(t;{\bm x}(t))$ samples. The fact that the structure seen after multiple samples of ${\bm x}(t)$ and $\hat{{\bm x}}(t;{\bm x}(t))$ resembles the target distribution -- though not yet scaled -- suggests that all the essential information about the target distribution emerges early in the dynamics. We see that the evolution of multiple $\hat{{\bm x}}(t;{\bm x}(t))$ samples precedes the changes observed in ${\bm x}(t)$. This observation is valuable as it suggests a method to reduce the number of steps required to generate samples from the target distribution.

\item We observe that the transition time’s dependence on $\beta$ is non-monotonic: there exists an optimal value of $\beta > 0$ where the transition time is minimized, and beyond this point, the transition time increases with $\beta$. 

\item Fig.~\ref{fig:particles_trajectory}, which displays exemplary trajectories, reveals a significant difference between the dynamics in the state space and the weighted state space. While the dynamics in the state space is relatively direct -- progressing from the origin to the target sample -- the dynamics in the weighted state space is much more complex. We observe that $\hat{{\bm x}}(t;{\bm x}(t))$ meanders between different "species" (associated with various terms in the Gaussian mixture) before locking onto the one that will eventually become the final sample from the target distribution.

\end{enumerate}

The UHIS algorithm should be compared with other general purpose sampling algorithms. (General purpose means going beyond comparing with the special algorithms tuned to the use case, like one we use to bench mark in the case of the sampling from the Gaussian mixture.) We have two types of such general samplers to compare UHIS with -- a Simulated Annealing (SA) Markov Chain Monte Carlo (MCMC) algorithm \cite{kirkpatrick_optimization_1983} and the Path Integral Sampler (PIS) of \cite{zhang_path_2022}. 
	
When comparing UHIS and SA-MCMC, we first of all note that both methods stem from the same class but exhibit important distinctions. SA-MCMC typically begins with an unbiased, uniformly distributed initialization across the domain, which is gradually adjusted through a slow, adiabatic process to converge to the target distribution. In contrast, UHIS starts from a $\delta$-function and does not require gradual adjustment. Notably, UHIS aims to rapidly optimize the sampling process, achieving i.i.d. samples from the target distribution in as few steps as possible.

The relationship between PIS and UHIS is even closer, as both methods are rooted in the Gaussian-Green function paradigm of Stochastic Optimal Control (SOC), which is a central theme of this manuscript and, consequently, to UHIS. Specifically, both approaches are connected to the $V = 0$ formulation, as derived in \cite{tzen_theoretical_2019} and reproduced in Section \ref{sec:V=0}. However, \cite{zhang_path_2022} takes a different route by approximating the optimal control using a NN. Instead of directly fitting the explicit solution from Eq.~(\ref{eq:Rag5}), they employ an NN-based approach akin to Model Predictive Control (MPC), where the NN is trained by solving Eq.~(\ref{eq:SOC}) with $V = 0$ and the control term ${\bm u}$ replaced by the NN. As a result, PIS becomes significantly more computationally intensive than UHIS, both during the training phase and in inference. Notably, a thorough evaluation of PIS performance on the well-known Gaussian mixture problem over a $3 \times 3$ grid was reported in \cite{richter_improved_2024}, where a particularly challenging case for PIS was identified. We replicated this challenging scenario and demonstrated that UHIS successfully resolves the case that was problematic for PIS, as shown in Figs.~\ref{fig:cl2},\ref{fig:boxplot}. Note that a significant aspect of \cite{richter_improved_2024}'s critique of PIS lies in its extreme sensitivity to the choice of the loss function used for training the NN. By contrast, UHIS avoids this issue entirely, as it does not rely on NN training and thus the selection of a specific loss function. 
	
We would also like to include a disclaimer: by focusing on UHIS, which is free from NNs, we are not dismissing the potential of NN-based schemes, and likely hybrid approaches, as the ultimate solution to the challenge of sampling from a target distribution represented by an energy function. In fact, we believe that a modification of the PIS approach, which leverages the Gaussian structure of the underlying Green functions -- such as the one expressed in the case of $V=0$ in Eq.~(\ref{eq:Rag5}) -- but still incorporates NNs, could be a promising direction for future research.
	
We conclude this section with some general remarks on other recently emerging approaches, such as the so-called Boltzmann generators \cite{noe_boltzmann_2019,midgley_flow_2023,zheng_towards_2023,blessing_beyond_2024,sendera_improved_2024,klein_transferable_2024}, which are alternative generative AI tools aimed at efficiently addressing the same challenge (of sampling from a target distribution represented by an energy function). Given that PIS, in its current implementation, has not fully exploited the integrable structure of the underlying SOC, it can be considered on par with Boltzmann generators. Therefore, and in light of our earlier comments on improving PIS, we are keenly anticipating further research that will advance successful hybrids of UHIS, PIS, and Boltzmann generators.
	
\section{Experiments: Empirical Sampling based on Ground Truth Data}\label{sec:GT-sampling}

	\begin{figure}[h!]
		\subfigure[]{
			\centering
			\includegraphics[scale=0.33]{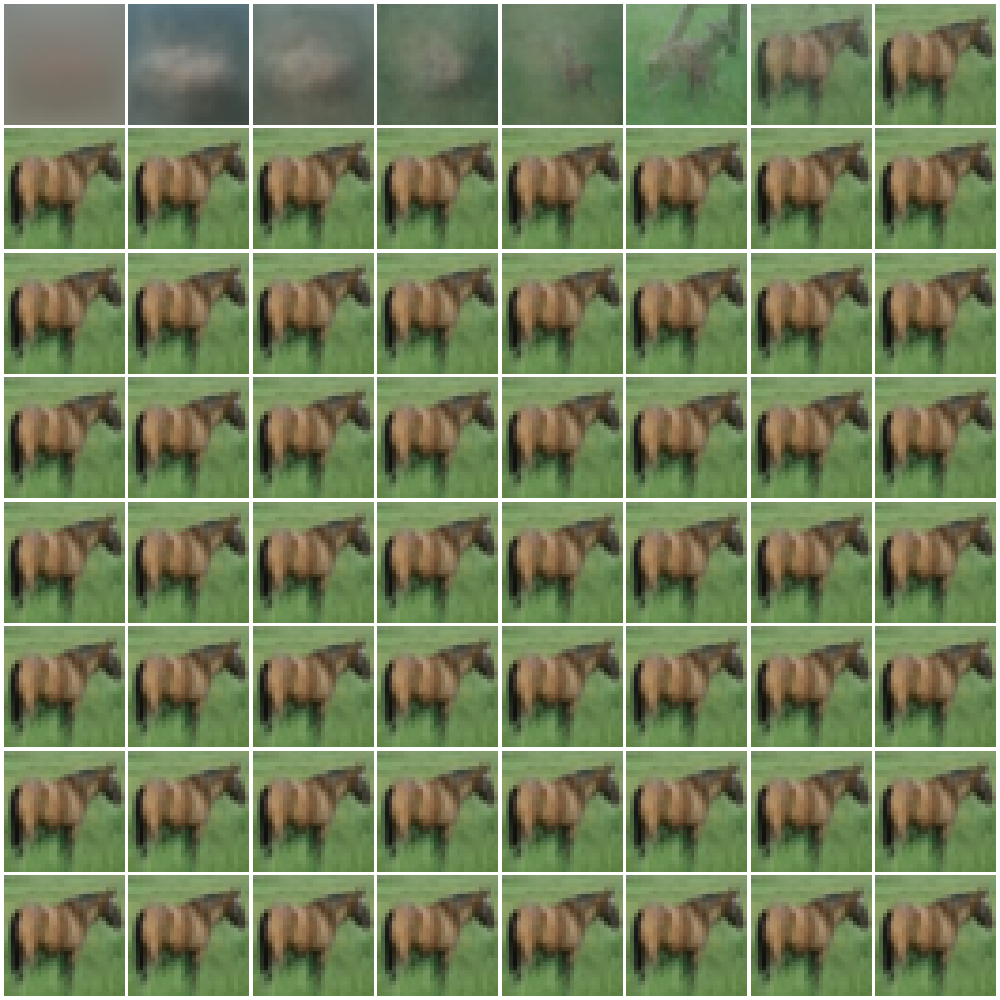} }
		\subfigure[]{
			\centering
			\includegraphics[scale=0.33]{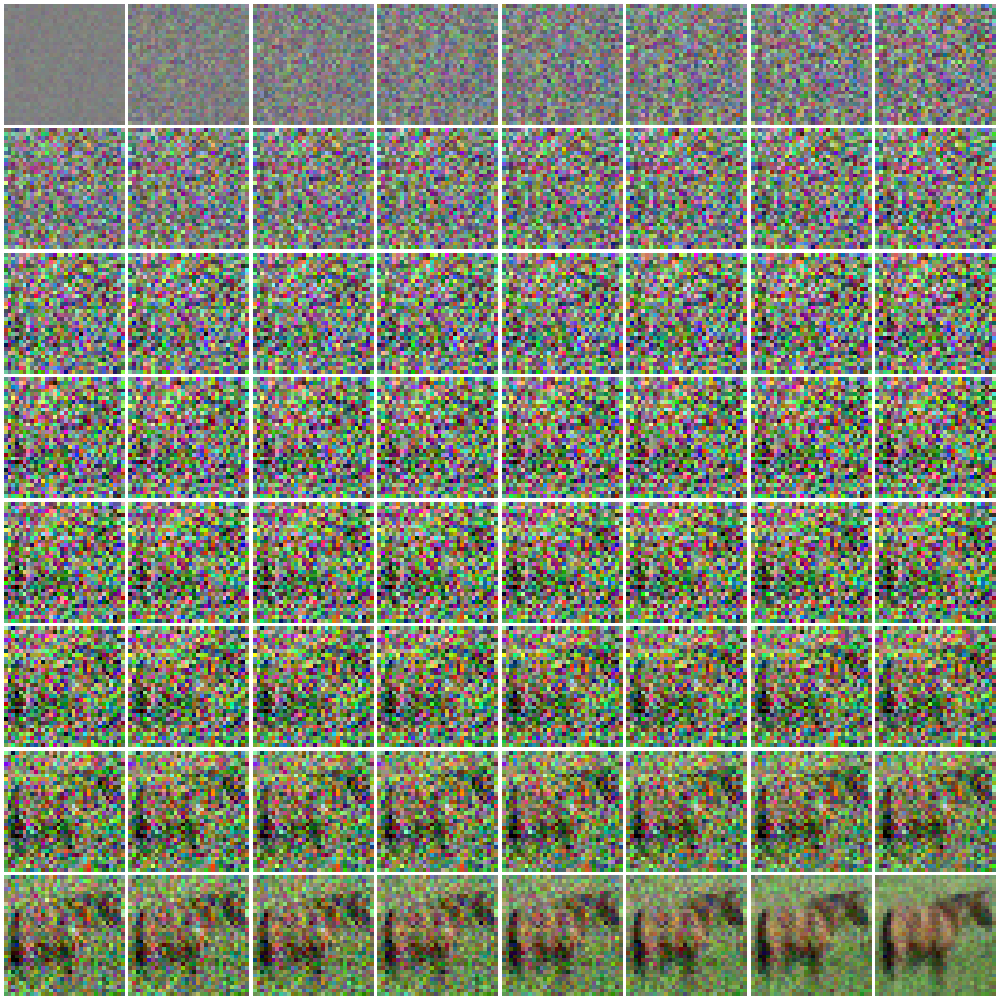} }
		\vspace{-5pt}
		\caption{Sample evolution from $0 \to 1$ : (left) the current weighted state, $\hat{\bm x}(t;{\bm x}(t))\doteq \sum_s\bm{y}^{(s)} w(\bm{y}^{(s)}|t;\bm{x}(t))$, and (right) the current state, $\bm{x}(t)$. 
  } 
		\label{fig:sample_evl} 
	\end{figure}

 \begin{figure}[h!]
		\subfigure[]{
			\centering
			\includegraphics[scale=0.4]{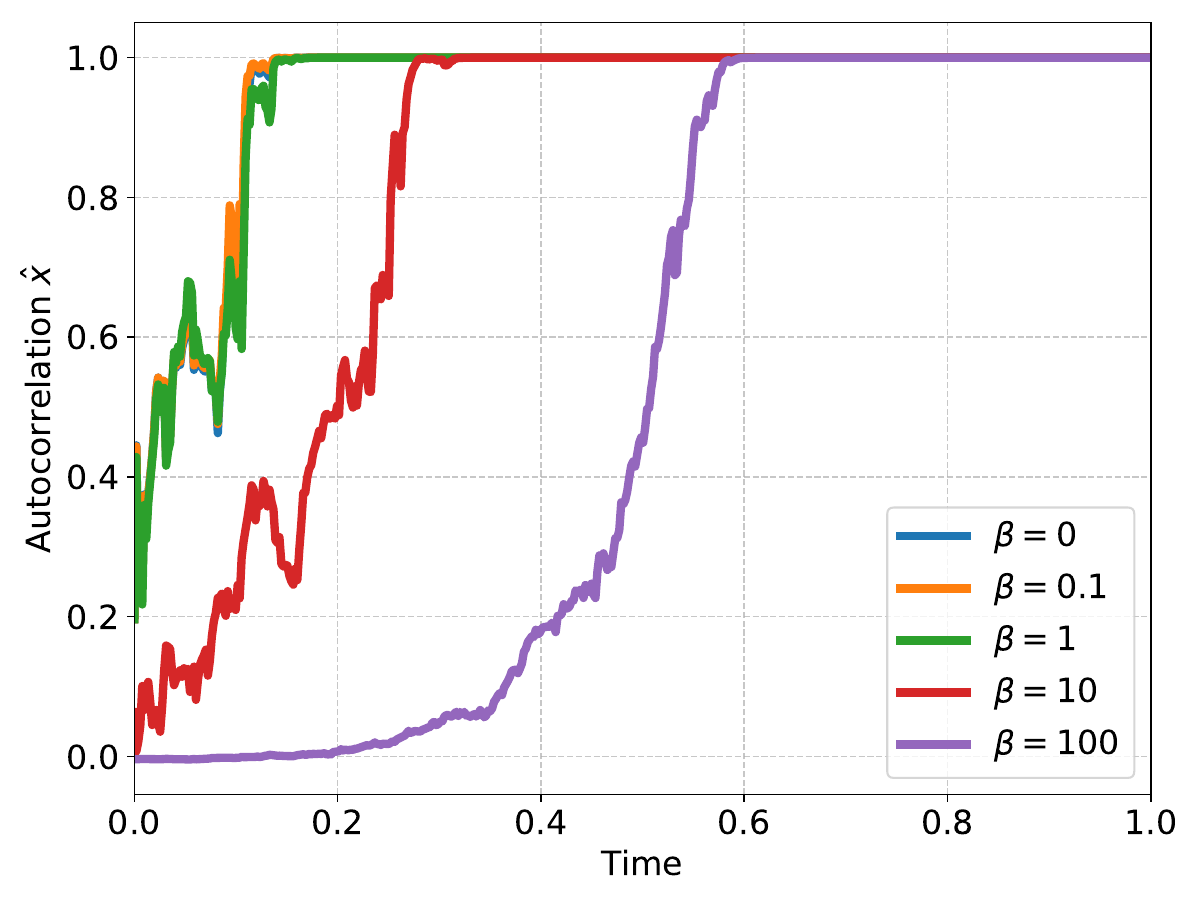} }
		\subfigure[]{
			\centering
			\includegraphics[scale=0.4]{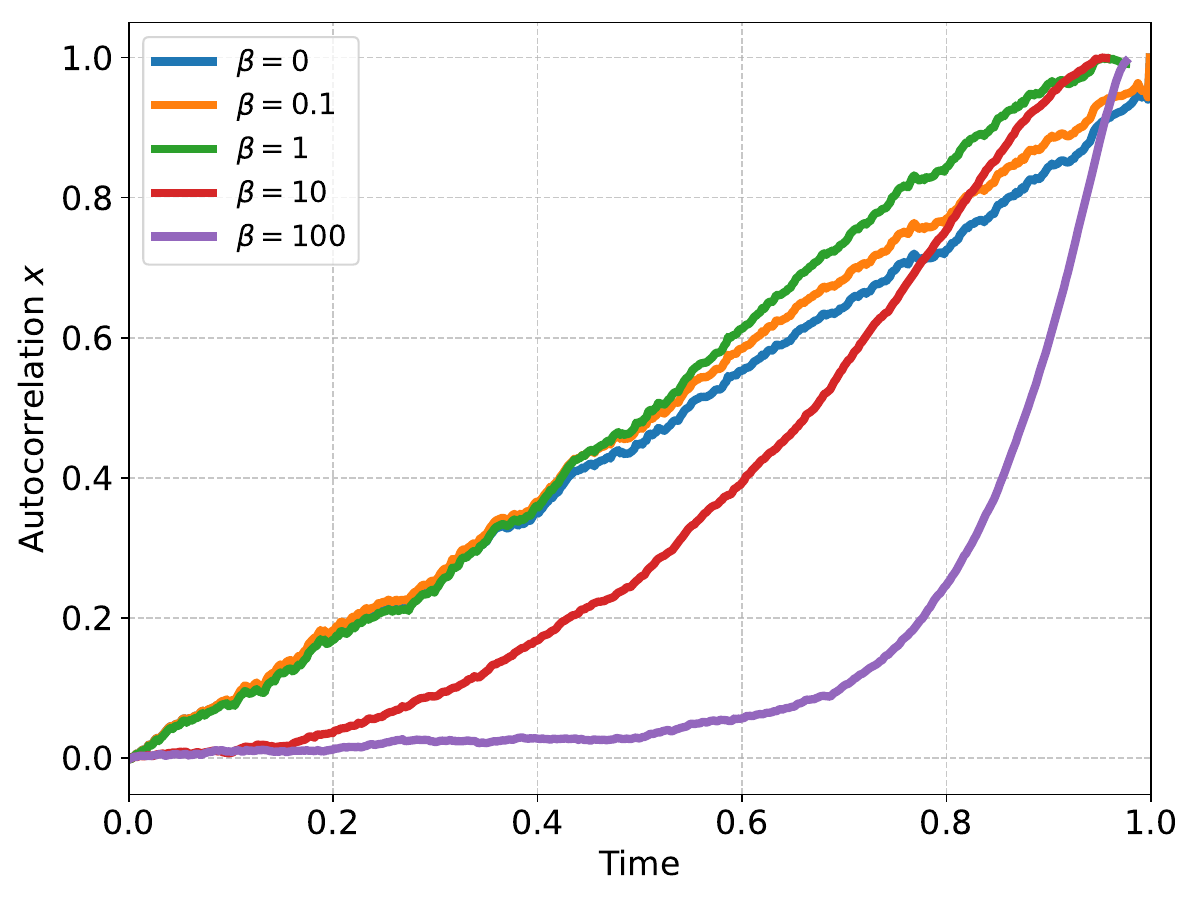} }
		\vspace{-5pt}
		\caption{Auto correlations for a single sample: (a-left) $(\hat{\bm x}^T(t;{\bm x}(t)){\bm x}(1))$ and (b-right) $(\bm{x}^T(t){\bm x}(1))$.
  } 
		\label{fig:sample_auto} 
	\end{figure}

The sample-based approximation of Eq.~(\ref{eq:gen-u*-via-target}) is given by:
	\begin{gather}\label{eq:u*-sample}
		u^*(t; {\bm x}) \approx \nabla_{\bm x} \log\left(\frac{1}{S} \sum_{s=1}^S 
		\frac{G_-(t; {\bm x}; \bm{y}^{(s)})}{G_+(1; \bm{y}^{(s)}; {\bm 0})}\right).
	\end{gather}
In this section, as in the previous one, we focus on the case where ${\bm A} = {\bm f} = 0$ and $V(t; {\bm x}) = \beta |{\bm x}|^2/2$. Under these conditions, and consistent with Eqs.~(\ref{eq:IS-u},\ref{eq:hat-x}), Eq.~(\ref{eq:u*-sample}) simplifies to:
	\begin{gather}\label{eq:u*-sample-1}
		u^*(t; {\bm x}) \approx  
		\frac{\sqrt{\beta}}{\sinh((1-t)\sqrt{\beta})}\left(\hat{\bm x}(t;{\bm x}) - {\bm x}\cosh((1-t)\sqrt{\beta})\right),\quad \hat{\bm x}(t;{\bm x})\doteq \sum_s\bm{y}^{(s)} w(\bm{y}^{(s)}|t;\bm{x}),
	\end{gather}
with the (probability) weights $w(\bm{y}^{(s)}|t;\bm{x})$ defined as:
	\begin{gather}
		w(\bm{y}^{(s)}|t;\bm{x}) \doteq \frac{\frac{G_-(t; {\bm x}; \bm{y}^{(s)})}{G_+(1; \bm{y}^{(s)}; {\bm 0})}}{\sum_{s'}^S \frac{G_-(t; {\bm x}; {\bm x}^{(s')})}{G_+(1; {\bm x}^{(s')}; {\bm 0})}} = \text{softmax}\left( \log \left( \frac{G_-(t; {\bm x}; \bm{y}^{(s)})}{G_+(1; \bm{y}^{(s)}; {\bm 0})} \right) \right),
	\end{gather}
where the respective Green's functions are provided in Eqs.~(\ref{eq:FA-GF-}, \ref{eq:FA-GF+}); and $\hat{\bm x}(t;{\bm x})$ is the weighted current state, already familiar from the previous section. We generated a collection of samples from CIFAR-10 \footnote{https://www.cs.toronto.edu/~kriz/cifar.html}, computed the optimal control using Eq.~(\ref{eq:u*-sample-1}), and simulated the stochastic dynamics according to Eq.~(\ref{eq:SODE}). The results for the case $\beta = 0$ are shown in Fig.~\ref{fig:sample_evl} and similar figures for other values of $\beta$ are shown in Appendix \ref{app:figs}.

Upon examining the output at $t=1$, we observe the eventual emergence of one of the original Ground Truth (GT) samples. By design, the scheme tends to memorize rather than generalize, leading to the reproduction of one of the initial samples. Obviously, this behavior is not a desirable outcome in Generative AI. For true generalization -- i.e., generating new, legitimate samples -- simply passing information through the GT samples is insufficient. Additional information can be introduced in the form of symmetries or constraints specific to images. For instance, a probabilistic constraint that components of ${\bm x}$ corresponding to neighboring pixels are likely to be similar is a common approach used in image de-noising. Another option is to represent either ${\bm u}^{(*)}(t;{\bm x})$ or $\hat{\bm{x}}(t;{\bm x})$ via NNs with architectures optimized for image representation, such as U-Net or transformer models.
	
In future work, we plan to incorporate these constraints into the PID framework. However, in this manuscript, our focus remains on analyzing the dynamics of sample emergence as a result of optimal PID. Therefore, and assuming that the generalization (i.e., the ability of the diffusion scheme to generate new legitimate samples) and the analysis of sample emergence (whether memorized or new) are distinct questions, we focus in this Section on the latter -- analyzing the dynamics of how a sample emerges.

To elucidate our understanding of the temporal evolution of the current state, ${\bm x}(t)$, and the current weighted state, $\hat{\bm x}(t;{\bm x})$, we have also studied the auto-correlations between the current and final values --  $({\bm x}^T(t) {\bm x}(1))$ and $(\hat{\bm x}^T(t;{\bm x}(t)) {\bm x}(1))$  -- as the functions of time. The results are shown in Fig.~\ref{fig:sample_auto}.

The conclusions we draw from Figs.~\ref{fig:sample_evl}, \ref{fig:sample_auto} and the figures in Appendix \ref{app:figs}, which are notably consistent with the discussion in the previous section, are as follows:
\begin{enumerate}
    \item As shown in Fig.~\ref{fig:sample_auto}, and consistent with Fig.~\ref{fig:sample_evl}, the temporal evolution of a sample can be divided into two distinct phases: an initial phase where the signal grows from zero to a structure resembling the final sample, followed by a "fine-tuning" phase that produces a recognizable image. While the discussion in the previous section of sampling from the target distribution drew an analogy with the dynamic speciation transition in \cite{biroli_dynamical_2024}, the transition observed in Fig.~\ref{fig:sample_evl} appears more closely linked to the memorization/collapse transition discussed in \cite{behjoo_u-turn_2023} and \cite{biroli_dynamical_2024}.
    
    \item  Tracking the evolution of $\hat{\bm x}(t; {\bm x}(t))$, as depicted in the left panels of Figs.~\ref{fig:sample_evl}, \ref{fig:sample_auto}, reveals that a nearly complete image becomes visible in $\hat{\bm x}(t; {\bm x})$ early—upon the completion of the initial phase. If we were to output $\hat{\bm x}(t; {\bm x}(t))$ instead of ${\bm x}(t)$, the final image would be accessible much earlier in the process. In other words, and borrowing terminology from statistical physics, $\hat{\bm x}(t;{\bm x}(t))$ acts as the \textbf{order parameter} of the dynamic phase transition, signaling at an earlier stage that the sample generation process is nearing completion.

    \item Concerning the time dependence of the dynamic transition on $\beta$, examining the auto-correlations reported in Fig.~\ref{fig:sample_auto} shows that while the proper structure, as seen earlier in $\hat{\bm x}(t;{\bm x}(t))$, emerges at almost the same transition time for $\beta=0$, $\beta=0.1$, and $\beta=1$, the choice of $\beta=1$ is preferable from the perspective of auto-correlations in ${\bm x}(t)$. Further increasing $\beta$ results in a delayed transition. This is expected, as for larger $\beta$, the optimality of control -- central to our approach -- favors delaying the movement of ${\bm x}(t)$ away from the origin.
    
    \item Examining the dynamics during the early phase shows that $\hat{\bm x}(t;{\bm x}(t))$ does not immediately resemble the structure of the eventual final sample; instead, it takes time to explore and meander through other possible structures associated with other GT images. This is most clearly observed in Fig.~\ref{fig:sample_evl_10} for $\beta=10$, where transient images emerge in the second row of the left panel.
\end{enumerate}

\section{Related Recent Developments}\label{sec:recent}

\subsection{Related Recent Developments in the Harmonic Schrödinger Bridge Problem}\label{sec:prior-pubs}
	
	After obtaining the initial theoretical results presented in this manuscript, we came across closely related methodology reported in \cite{teter2024schr, teter2024weyl}. These papers propose a theory that, while yielding equivalent results, follows a different approach from ours. Both address the same fundamental question: finding the optimal control that maps one probability distribution into another under a fixed time horizon, with a cost-to-go function featuring a quadratic positive-definite term in the state vector, along with a quadratic contribution in the control.
	
	The approach in \cite{teter2024schr, teter2024weyl} extends the Schrödinger bridge methodology by formulating and solving equations for the harmonic and co-harmonic factors (also known as Schrödinger factors), as discussed in \cite{chen_optimal_2014, chen_stochastic_2021}. For a comprehensive historical discussion, see also \cite{chen_stochastic_2021}. In contrast, our approach begins with path-integral control \cite{kappen_path_2005}, using a fixed initial condition for the state and an additional terminal cost. After solving this auxiliary problem, we apply a dual transformation with respect to the terminal cost, allowing us to solve a Harmonic Schrödinger Bridge (HSB) problem. However, this approach maps a deterministic distribution (a $\delta$-function) to an arbitrary target distribution rather than between two arbitrary distributions.
	
	Notably, the general problem of constructing the harmonic Schrödinger map between two arbitrary probability distributions is intractable, even in the special case of $V=0$, as discussed in \cite{chen_optimal_2014, chen_stochastic_2021}. To address this challenge, the authors of \cite{teter2024schr, teter2024weyl} introduced an iterative Sinkhorn recursion that alternates between solving Schrödinger-like equations for the harmonic and co-harmonic factors. These factors can be interpreted as representing quantum mechanics in imaginary time within a quadratic potential, analogous to the quantum mechanics of a harmonic oscillator.
	
	Our problem, which involves growing an arbitrary probability distribution from a $\delta$-function (i.e., a deterministic initial condition), avoids these complications. We present the solution for this HSB-from-zero problem in closed form and demonstrate how to use this exact formula to sample from an arbitrary target probability distribution.
	
	It is also relevant to mention \cite{liu_generalized_2023}, which addresses a problem related to the HSB. Specifically, Lemma 3 in \cite{liu_generalized_2023} discusses mapping between two $\delta$-functions, presenting it as an example of exactly solvable path-integral control \cite{kappen_path_2005}. However, \cite{liu_generalized_2023} does not utilize the exact theory for algorithms or experiments but instead focuses on developing a generalized Schrödinger bridge for an arbitrary potential $V$.

\subsection{Föllmer Process and Interpolants}

The results presented in our manuscript can also be viewed as a generalization of the so-called \textbf{Föllmer Process} \cite{follmer_time_1986} -- a framework that minimizes the relative entropy with respect to the Wiener process, constructing a probabilistic path from a deterministic initial state to a target distribution over time. In this context, it is relevant to mention the recent developments in \cite{chen_probabilistic_2024}, which focus on \textbf{probabilistic forecasting}. The authors introduce the \textbf{stochastic interpolant}—an artificial dynamic governed by a stochastic differential equation (SDE), where the optimal drift term is learned via quadratic regression on available time-series data. This stochastic interpolant approach extends the Föllmer Process by providing a flexible means to manage uncertainties in forecasting through tuning the diffusion term in the SDE.

The approach in \cite{chen_probabilistic_2024}, based on optimizing a physically meaningful objective, is closely related to our method but also complementary in two key aspects. First, the setup in \cite{chen_probabilistic_2024} incorporates additional information in the design of the stochastic process; beyond the initial (deterministic) position and target distribution, it also aligns the resulting stochastic dynamics with a physically relevant time series within the interval. Second, the degrees of freedom employed in \cite{chen_probabilistic_2024} involve controlling the diffusion side, specifically through space- and time-dependent adjustments to the diffusion coefficient.

Given the overlap in goals—both approaches aim to generate accurate samples from a target distribution through a controlled stochastic process -- we foresee a promising synergy. In particular, our H-PID approach could be further generalized by incorporating adjustable diffusion coefficients, along with a corresponding adjustment to the quadratic control cost term, to retain the integrable Hopf-Cole-friendly structure of our scheme.

\subsection{Iterative Denoising Energy Matching}

The Iterative Denoising Energy Matching (iDEM) framework \cite{akhound-sadegh_iterated_2024} offers a compelling approach by employing a forward-backward SDE process to estimate the score function in a setting similar to that discussed in this manuscript—sampling from a target distribution defined by an energy function (hence, energy matching). This approach aligns closely with our H-PID, as iDEM essentially represents a particular case of H-PID with zero potential, $V = \beta = 0$, making it equivalent to the special case studied in \cite{tzen_theoretical_2019} and linking it to the Föllmer process \cite{follmer_time_1986}. Specifically, the iDEM methodology presented in \cite{akhound-sadegh_iterated_2024} introduces an analytical formulation for the score function as an integral over Gaussian noise, facilitating sampling from an unnormalized density function.

Interestingly, the connections between these approaches extend even further. In \cite{akhound-sadegh_iterated_2024}, the integral for the score function is evaluated numerically using a technique similar in spirit to the Universal Importance Sampling (UIS) approach proposed in this manuscript—that is, by generating samples from a Gaussian distribution independent of the target distribution. This insight reveals new connections in the literature and supports the analysis of H-PID in a broader setting, with varying $\beta$, more general Gaussian (and beyond) potentials $V$, as well as more sophisticated force and gauge functions, ${\bm A},{\bm f}\neq 0$, and diffusion generalizations of H-PID discussed in \cite{chen_probabilistic_2024}.

Furthermore, we anticipate exciting advancements through an optimal balance between NN-based representations, as explored in PIS \cite{zhang_path_2022} and \cite{chen_probabilistic_2024,hua_simulation-free_2024}, and UIS-based representations, as in \cite{akhound-sadegh_iterated_2024} and this manuscript, for the score/control functions.

\section{Conclusions \& Path Forward}\label{sec:conclusions}
	
The main high-level finding of this manuscript builds on a long history of work in stochastic optimal control and non-equilibrium statistical mechanics. We have discovered that there exists a large and expressive family of stochastic optimal control formulations of the \textbf{Path Integral Control} type \cite{kappen_path_2005}, which allows for exact solutions to the problem of generating i.i.d. samples from complex multivariate probability distributions, represented either through an energy function or through samples. By "exact," we mean that there is an explicit expression for the optimal control (or equivalently, the score function in generative AI terminology) within the underlying stochastic differential equation of the Drift-\textbf{Diffusion} type. This equation describes the continuous evolution from an initial state at $t = 0$ to a target probability distribution at $t = 1$. Specifically, the optimal control is expressed as the gradient of the logarithm of a convolution between the target probability distribution and a ratio of two Green's functions, which describe linear evolution backward and forward in time in an auxiliary problem. While this statement applies to any PIC formulation, in the special cases where the Green's functions are Gaussian, this exactly solvable auxiliary problem becomes reminiscent of quantum mechanics (in imaginary time) within a \textbf{Harmonic} potential.

We have applied this theoretical insight to develop new algorithms that we believe can become competitive with state-of-the-art techniques in generative AI. Importantly, our approach does not rely on NNs, thereby avoiding the black-box nature of underlying functions and the lengthy, resource-intensive training typically required for such models. Instead, our methods can be implemented (and in fact prefers) using solely CPU computations. In the case where the target probability distribution is represented by an energy function, the score function (optimal control function) is computed using a universal Importance Sampling algorithm, independent of the energy function. In the other case, where the target probability distribution is represented by GT samples, the score function is expressed as a sum over GT samples.

Interested in understanding and improving the newly suggested PID methodology, we conducted a detailed analysis of the H-PID algorithm on the case which is controlled by a single parameter, $\beta$, that measures the relative strength of the quadratic potential term in the control cost. Our experiments with the two aforementioned algorithms reveal several intriguing observations: The analytic transparency of our approach allowed us to uncover important hidden structures in the optimal diffusion process and its dependence on $\beta$. This structure was discovered through the analysis of the {\bf current weighted state}, $\hat{\bm x}(t; {\bm x}(t))$, which depends on the current state, ${\bm x}(t)$, of the optimal stochastic process observed at time $t$. The temporal evolution of the samples, observed via the current weighted state, reveals a dynamic phase transition characterized by two distinct stages: an initial phase of growth followed by a fine-tuning phase that leads to the final sample. The analysis of auto-correlations between current and final values shows that essential structural information is present early in the process -- but hidden, encoded in the weighted current state -- with the choice of $\beta$ influencing the timing and progression of the transition. We conclude that $\hat{\bm x}(t;{\bm x}(t))$ acts as the \textbf{order parameter} of the dynamic phase transition. This analysis provides valuable insights for optimizing the sampling process in the future.

Looking ahead, we envision further theoretical development, algorithmic refinement, and experimentation. On the practical side, we see several extensions of the integrable SOC approach to more specialized applications, such as: (a) fine-tuning pre-trained models with additional personalized biases expressed as corrections to the energy function (see, for example, \cite{domingo-enrich_adjoint_2024}); (b) mixing multiple pre-trained models for enhanced performance; (c) biasing sampling towards extreme events, where conventional AI tools struggle to generalize; and (d) extending to the cases where the two ways to represent the target distributions (via energy function and via sample) are combined. 
	
We are eager to explore ways to enhance the {\bf Harmonic Path Integral Diffusion} approach by: (a) developing {\bf Forced} (non-conservative) use cases with non-zero gauge potential and force; (c) addressing the challenge of GT sample memorization, both with and without NNs (as discussed in Section \ref{sec:GT-sampling}); and (d) implementing a non-universal importance sampling (IS) algorithm, which explicitly depends on the energy function, as outlined in Section \ref{sec:IS-nonuniversal}.
	
Furthermore, we plan to integrate the tools introduced in this manuscript with other generative model approaches. In particular, we aim to gain a deeper understanding of how information is lost, acquired, or mixed during diffusion processes in AI, as seen in \cite{behjoo_u-turn_2023}, which can lead to dynamic speciation and collapse/memorization phase transitions \cite{biroli_dynamical_2024}. Additionally, we hope to explore how spatio-temporal mixing, as discussed in \cite{behjoo_space-time_2024}, can be incorporated into the approach of this manuscript.  

\section*{Acknowledgments}
	
We would like to express our gratitude to Vladimir Chernyak for numerous insightful discussions during the early stages of this project, particularly for encouraging us to explore generalizations of Green's function integrability to forced/non-conservative cases (${\bm A}, {\bm f} \neq 0$). We are also thankful to Sungsoo Ahn for  bringing to our attention the latest research on sampling from a target distribution defined by an energy function and also to  Eric Vanden-Eijnden for attracting our attention to related line of work on generalized Fölmer processes.

	\appendix
	
	\section{Green Function for Uniform Quadratic Potential} \label{sec:GFuni}
	
	\subsection{Reverse Dynamics} \label{sec:GF-}
	
	We are looking for solution of Eq.~(\ref{eq:gen-GF-}) for $V=\beta{\bm x}^2/2$ in the Gaussian form
	\begin{gather}\label{eq:G-Gauss}
		G_-(t;{\bm x};{\bm y})=\exp\left(-\frac{A(t)}{2}{\bm x}^2+B(t)\left({\bm x}^T{\bm y}\right)-C(t)\right).
	\end{gather}
	This results in the following system of ODEs
	\begin{gather}\label{eq:ABC}
		\dot{A}+\beta=A^2,\quad \dot{B}=A B,\quad
		\dot{C}=\frac{B^2{\bm y}^2}{2}-\frac{A}{2},
	\end{gather}
	which should also support the specific $t\to 1$ singular asymptotic
	\begin{gather}\label{eq:ABC-sing}
		t\to 1:\quad  A(t),B(t)\to \frac{1}{1-t},\quad C(t)\to \frac{{\bm y}^2}{2(1-t)}+\frac{1}{2}\log\left(2\pi(1-t)\right).
	\end{gather}
	The solution of Eqs.~(\ref{eq:ABC},\ref{eq:ABC-sing}) is
	\begin{gather} \label{eq:ABC-sol}
		A(t)=\frac{\sqrt{\beta}}{\tanh\left((1-t)\sqrt{\beta}\right)},\quad B(t)=\frac{1}{\sinh\left((1-t)\sqrt{\beta}\right)},\quad e^{-C(t)}= \frac{\exp\left(-\frac{{\bm y}^2 \sqrt{\beta}}{2\tanh\left((1-t)\sqrt{\beta}\right)}\right)}{\sqrt{2\pi\sinh\left((1-t)\sqrt{\beta}\right)/\sqrt{\beta}}}.
	\end{gather}
	Summarizing,
	\begin{gather}\label{eq:GF}
		G_-(t;{\bm x};{\bm y})=\frac{\exp\left(-\sqrt{\beta}\frac{({\bm x}^2+{\bm y}^2)\cosh\left((1-t)\sqrt{\beta}\right)-2({\bm x}^T{\bm y})}{2\sinh\left((1-t)\sqrt{\beta}\right)}\right)}{\sqrt{2\pi\sinh\left((1-t)\sqrt{\beta}\right)/\sqrt{\beta}}}.
	\end{gather}
	
	\subsection{Forward Dynamics} \label{sec:GF+}
	
	In this Appendix we construct solution of Eq.~(\ref{eq:gen-GF+}) for $V=\beta{\bm x}^2/2$. Logic of the derivation is identical to the we follow in the previous Subsection. We therefore present here results (skipping discussions):
	\begin{align}\label{eq:G+}
		& G_+(t;{\bm x};{\bm y}) =\exp\left(-\frac{A_+(t)}{2}{\bm x}^2+B_+(t)\left({\bm x}^T{\bm y}\right)-C_+(t)\right),\\
		\label{eq:ABC+}
		& -\dot{A}_++\beta=A_+^2,\quad -\dot{B}_+=A_+ B_+,\quad
		-\dot{C}_+=\frac{B^2_+{\bm y}^2}{2}-\frac{A_+}{2},\\
		& \label{eq:G+asymp}
		t\to 0:\quad  A_+(t),B(t)\to \frac{1}{t},\quad C_+(t)\to \frac{{\bm y}^2}{2t}+\frac{1}{2}\log\left(2\pi t\right),\\ \label{eq:ABC+2}
		&    A_+(t)=\frac{\sqrt{\beta}}{\tanh\left(t\sqrt{\beta}\right)},\quad B_+(t)=\frac{1}{\sinh\left(t\sqrt{\beta}\right)},\quad e^{-C(t)}= \frac{\exp\left(-\frac{{\bm y}^2 \sqrt{\beta}}{2\tanh\left(t\sqrt{\beta}\right)}\right)}{\sqrt{2\pi\sinh\left(t\sqrt{\beta}\right)/\sqrt{\beta}}},\\ \label{eq:GF+2} &
		G_+(t;{\bm x};{\bm y})=\frac{\exp\left(-\sqrt{\beta}\frac{({\bm x}^2+{\bm y}^2)\cosh\left(t\sqrt{\beta}\right)-2({\bm x}^T{\bm y})}{2\sinh\left(t\sqrt{\beta}\right)}\right)}{\sqrt{2\pi\sinh\left(t\sqrt{\beta}\right)/\sqrt{\beta}}}.
	\end{align}
	
	\subsection{Universal Stationary Point}\label{app:uni-IS}

Here we focus on computing ${\bm y}_*(t;{\bm x})$ and $\hat{\bm H}(t;{\bm x})$ defined implicitly in Eq.~(\ref{eq:y*}) and Eq.~(\ref{eq:H}), respectively.

Utilizing Eqs.~(\ref{eq:GF},\ref{eq:GF+2}) we derive 
\begin{align}\label{eq:G-ratio}
	& \frac{G_-(t;{\bm x};{\bm y})}{G_+(1;{\bm y};{\bm 0})}= 
	\sqrt{\frac{\sinh\left(\sqrt{\beta}\right)}{\sinh\left((1-t)\sqrt{\beta}\right)}}
	\exp\left(-\frac{\sqrt{\beta}}{2}
	\left(({\bm x}^2+{\bm y}^2)\coth\left((1-t)\sqrt{\beta}\right)-{\bm y}^2\coth\left(\sqrt{\beta}\right)-\frac{2({\bm x}^T{\bm y})}{\sinh\left((1-t)\sqrt{\beta}\right)}\right)
	\right),\\
	& \nabla_{\bm y}\log\left(\frac{G_-(t;{\bm x};{\bm y})}{G_+(1;{\bm y};{\bm 0})}\right)= 
	\sqrt{\beta} \left({\bm y} \left(\coth (\sqrt{\beta})- \coth \left((1-t)\sqrt{\beta}\right)\right)+{\bm x}\frac{1}{\sinh ((1-t)\sqrt{\beta})}\right),  \label{eq:y*-1}\\
	& H_{ij}=-\partial_{y_i}\partial_{y_j}\log\left(\frac{G_-(t;{\bm x};{\bm y})}{G_+(1;{\bm y};{\bm 0})}\right)\Bigg|_{{\bm y}\to {\bm y}_*}=\delta_{ij}\sqrt{\beta} \left(\coth \left((1-t)\sqrt{\beta}\right)-\coth (\sqrt{\beta})\right),
	\label{eq:H-1}
\end{align}
where the last expression gives us explicit -- isotropic and ${\bm x}$-independent -- expression for the Hessian.

Substituting Eq.~(\ref{eq:y*-1}) into Eq.~(\ref{eq:y*}) we derive explicit expression for ${\bm y}_*$ as a function of ${\bm x}$ and $t$
\begin{gather}\label{eq:y*-2}
	{\bm y}_*=\frac{{\bm x}}{\cosh \left((1-t)\sqrt{\beta}\right)-\sinh \left((1-t)\sqrt{\beta}\right)\coth(\sqrt{\beta})}.
\end{gather}

\subsection{Non-Universal -- Energy Function dependent -- Stationary Point}\label{app:non-uni-IS}

Substituting Eqs.~(\ref{eq:y*-1},\ref{eq:H-1}) into Eqs.~(\ref{eq:y-E},\ref{eq:H-E}) we derive
\begin{align}\label{eq:y-d}
	\nabla_{{\bm y}^\diamond} \tilde{E}({\bm y}^\diamond) & = {\bm x},\quad \tilde{E}({\bm y}^\diamond)\doteq \frac{E({\bm y}^\diamond)+|{\bm y}^\diamond|^2 \left(\coth \left((1-t)\sqrt{\beta}\right)-\coth (\sqrt{\beta})\right)\sqrt{\beta}/2}{\sinh ((1-t)\sqrt{\beta})\sqrt{\beta}},
	\\ 
	& H^\diamond_{ij}=\sinh ((1-t)\sqrt{\beta})\sqrt{\beta}\ \partial_{y^\diamond_i}\partial_{y_j^\diamond} \tilde{E}({\bm y}^\diamond).
	\label{eq:H-d}
\end{align}

\section{Green Function for General Quadratic Potential} \label{sec:GFgen}

\subsection{Reverse Dynamics} \label{sec:GF-gen}

As in other cases we are looking for Green function solving Eq.~(\ref{eq:gen-GF-}) in the case of $V={\bm x}^T\hat{\bm \beta}{\bm x}/2$ in the Gaussian form
\begin{gather}\label{eq:GF-gen}
	G_-(t;{\bm x};{\bm y})=\exp\left(-\frac{{\bm x}^T\hat{\bm A}(t) {\bm x}}{2}+\left({\bm x}^T\hat{\bm B}(t){\bm y}\right)-C(t)\right),
\end{gather}
where $\hat{\bm A}$ is symmetric and positive-definite. This results in the following system of ODEs
\begin{gather}\label{eq:ABC-gen}
	\frac{d}{dt}\hat{\bm A}+\hat{\bm \beta}=\hat{\bm A}^2,\quad \frac{d}{dt}\hat{\bm B}=\hat{\bm A} \hat{\bm B},\quad
	\frac{d}{dt} C=\frac{{\bm y}^T\hat{\bm B}^T\hat{\bm B}{\bm y}}{2}-\frac{\text{tr}[\hat{\bm A}]}{2},
\end{gather}
which should also support the specific $t\to 1$ singular asymptotic
\begin{gather}\label{eq:ABC-sing-gen}
	t\to 1:\quad  \hat{\bm A}(t),\ \hat{\bm B}(t)\to \frac{\hat{\bm I}}{1-t},\quad C(t)\to \frac{{\bm y}^2}{2(1-t)}+\frac{1}{2}\log\left(2\pi(1-t)\right).
\end{gather}
The solution of Eqs.~(\ref{eq:ABC-gen},\ref{eq:ABC-sing-gen}) is
\begin{gather}\label{eq:ABC-sol-gen}
	\hat{\bm A}(t)=\frac{\sqrt{\hat{\bm \beta}}}{\tanh\left((1-t)\sqrt{\hat{\bm \beta}}\right)},\quad \hat{\bm B}(t)=\left(\sinh\left((1-t)\sqrt{\hat{\bm \beta}}\right)\right)^{-1},\quad 
	e^{-C(t)}= 
	\frac{
		\exp\left(-\frac{1}{2} {\bm y}^T \frac{\sqrt{\hat{\bm \beta}}}{2\tanh\left((1-t)\sqrt{\hat{\bm \beta}}\right)}{\bm y}\right)
	}{
		\sqrt{(2\pi)^d\det\left(\sinh\left((1-t)\sqrt{\hat{\bm \beta}}\right)/\sqrt{\hat{\bm \beta}}\right)}
	}.
\end{gather}

\subsection{Forward Dynamics} \label{sec:GF+gen}

Logic of the derivation is identical to the one we follow in the previous Subsection. We therefore present here results (skipping discussions):
\begin{align}\label{eq:G+gen}
	& G_+(t;{\bm x};{\bm y}) =\exp\left(-\frac{1}{2}{\bm x}^T\hat{\bm A}_+(t){\bm x}+{\bm x}^T\hat{\bm B}_+(t){\bm y}-C_+(t)\right),\\
	\label{eq:ABC+gen}
	& -\frac{d}{dt}\hat{\bm A}_++\hat{\bm \beta}=\hat{\bm A}_+^2,\quad -\frac{d}{dt} \hat{\bm B}_+=\hat{\bm A}_+ \hat{\bm B}_+,\quad
	-\frac{d}{dt} C_+=\frac{{\bm y}^T \hat{\bm B}_+^T\hat{\bm B}_+{\bm y}}{2}-\frac{\text{tr}[\hat{\bm A}_+]}{2},\\
	& \label{eq:G+asymp-gen}
	t\to 0:\quad  \hat{\bm A}_+(t),\ \hat{\bm B}(t)\to \frac{\hat{I}}{t},\quad C_+(t)\to \frac{{\bm y}^2}{2t}+\frac{1}{2}\log\left(2\pi t\right),\\ \label{eq:ABC+2-gen}
	&    \hat{\bm A}_+(t)=\frac{\sqrt{\hat{\bm \beta}}}{\tanh\left(t\sqrt{\hat{\bm \beta}}\right)},\quad \hat{\bm B}_+(t)=\left(\sinh\left(t\sqrt{\hat{\bm \beta}}\right)\right)^{-1},\quad e^{-C(t)}= \frac{\exp\left(-{\bm y}^T\frac{\sqrt{\hat{\bm \beta}}}{2\tanh\left(t\sqrt{\hat{\bm \beta}}\right)}{\bm y}\right)}{\sqrt{(2\pi)^d\det\left(\sinh\left(t\sqrt{\hat{\bm \beta}}\right)/\sqrt{\hat{\bm \beta}}\right)}},
\end{align}
where $\hat{\bm A}$ is symmetric and positive definite.

\section{Auxiliary  Figures} \label{app:figs}
	\begin{figure}[h!]
		\centering
		\begin{adjustbox}{valign=c,clip,trim=0.75in 6in 0in 0in} 
			\includegraphics[scale=1]{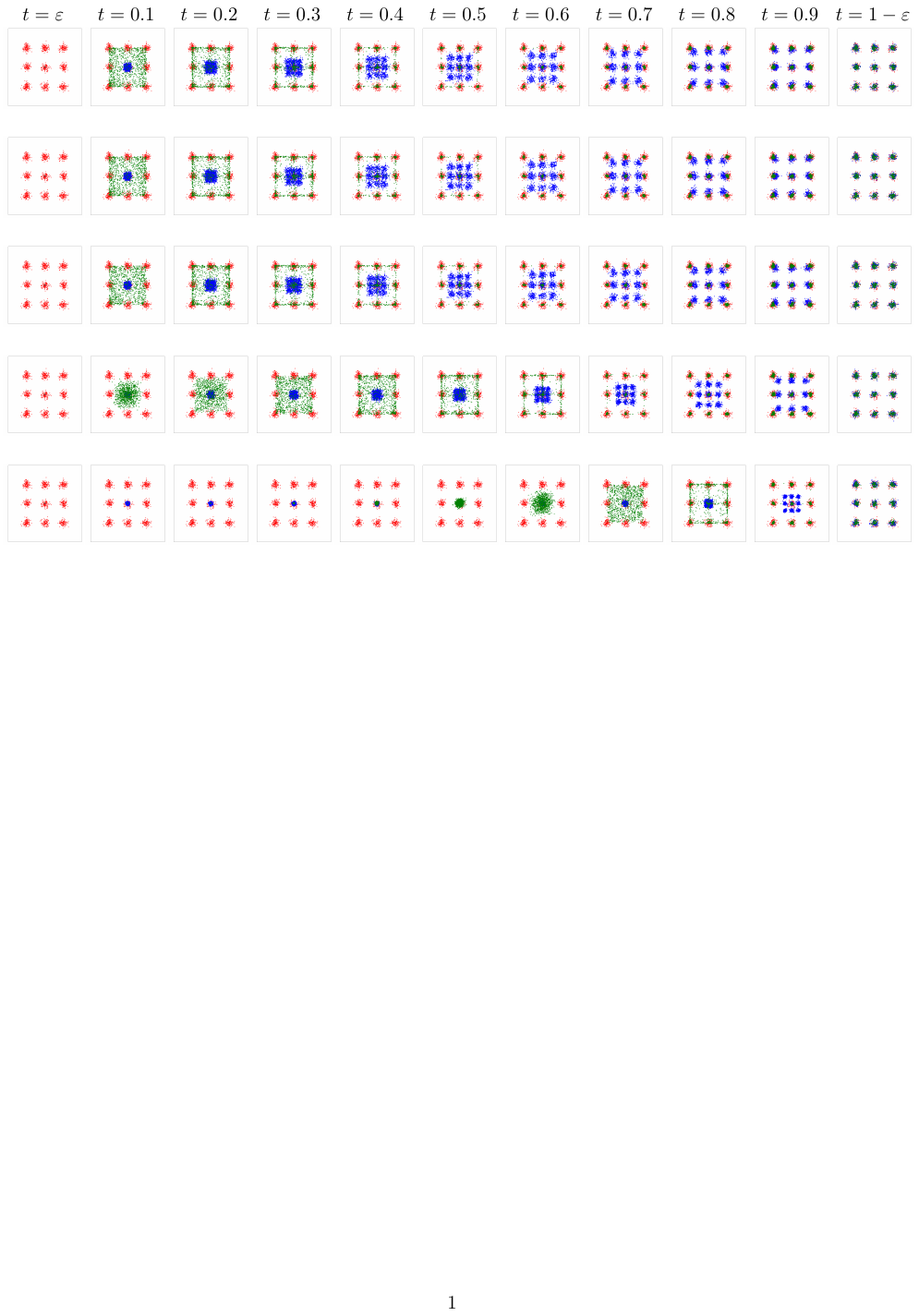}
		\end{adjustbox}
		\caption{ The red circles represent i.i.d. samples drawn from the target distribution, which is a mixture of nine Gaussian components arranged in a \((3 \times 3)\) square grid, where distance between nearest grid points is $5$, variance of the Gaussians is 0.5. The blue circles illustrate the temporal evolution of samples governed by Eq.~(\ref{eq:SODE}), where the time interval \( t = (0, 1) \) is discretized into 200 steps and the optimal control \({\bm u}(t; {\bm x}) \to {\bm u}^*(t; {\bm x})\) is defined according to Eq.~(\ref{eq:UHIS}) with \(N=10000\). Finally green dots show the evolution of $\hat{\bm x}(t; {\bm x})$. Each row corresponds to a different value of $\beta$ ($\beta = 0$, $\beta = 0.1$, $\beta = 1$, $\beta = 10$, and $\beta = 100$, respectively).
  }
		\label{fig:cl3}
	\end{figure}

\begin{figure}[h!]
	\subfigure[]{
		\centering
		\includegraphics[scale=0.33]{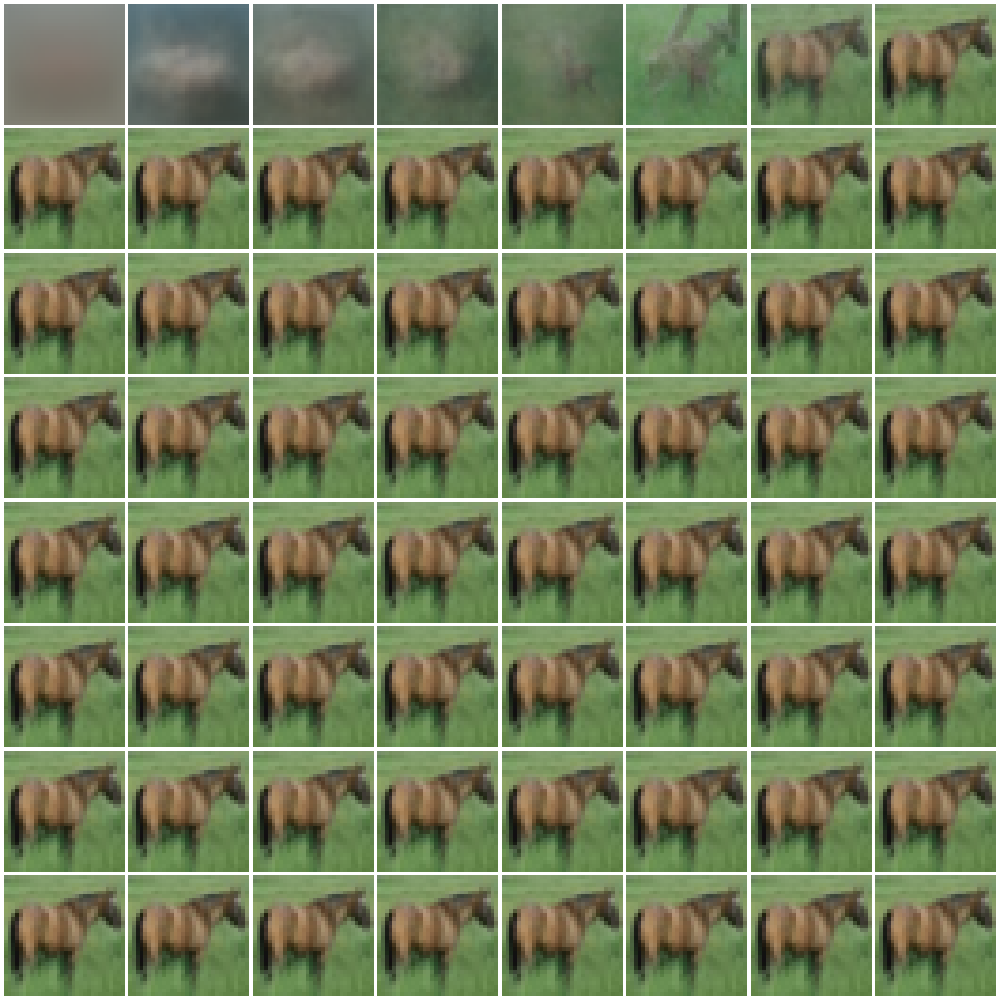} }
	\subfigure[]{
		\centering
		\includegraphics[scale=0.33]{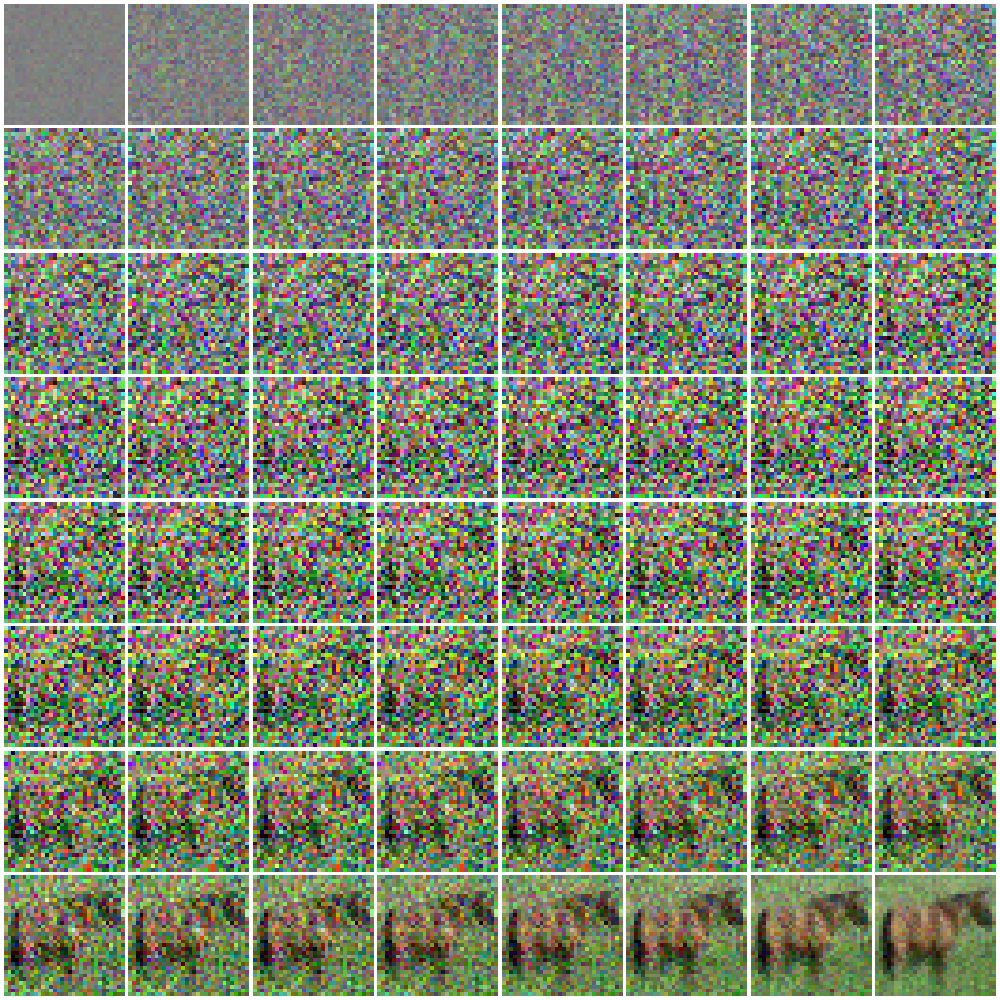} }
	\vspace{-5pt}
	\caption{Sample evolution from $0 \to 1$ : (Left) $\hat{\bm x}(t;{\bm x}(t))\doteq \sum_s\bm{y}^{(s)} w(\bm{y}^{(s)}|t;\bm{x}(t))$ and (Right) $\bm{x}(t)$. $\beta=0.1$} 
	\label{fig:sample_evl_0.1} 
\end{figure}

\begin{figure}[h!]
	\subfigure[]{
		\centering
		\includegraphics[scale=0.33]{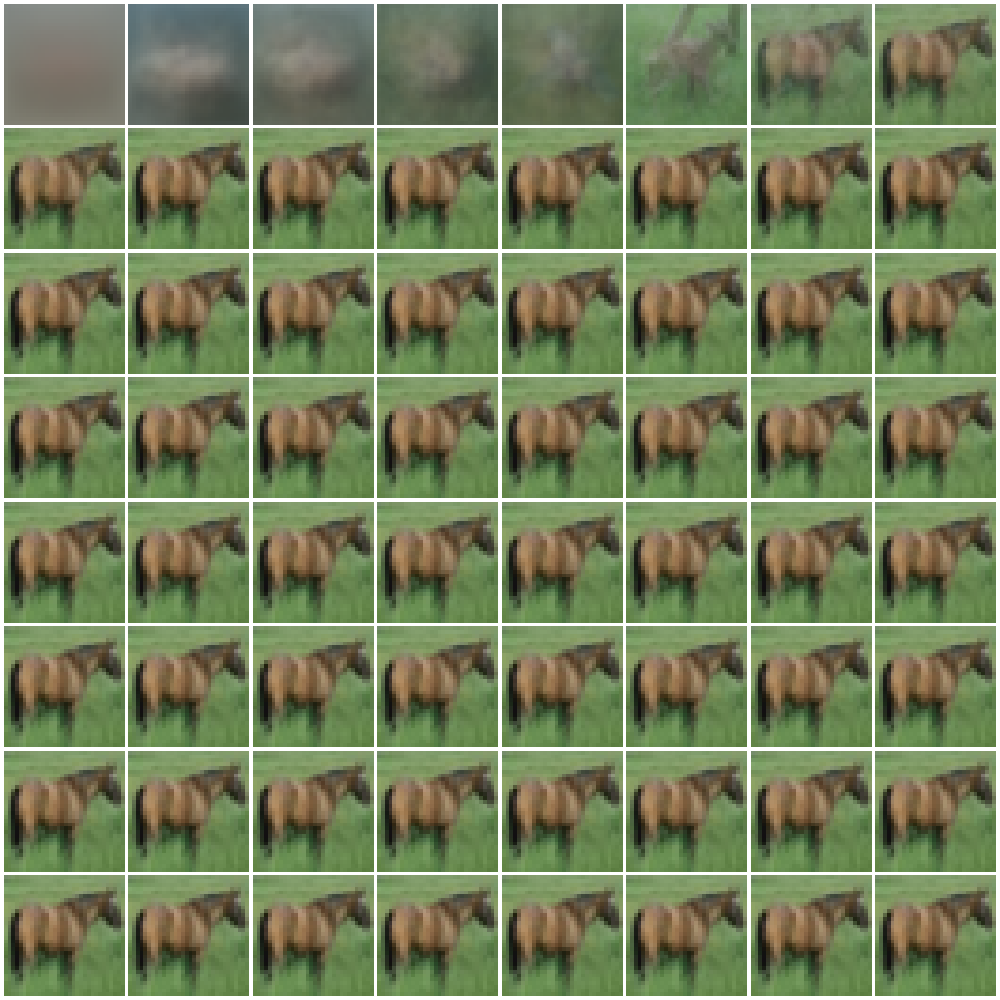} }
	\subfigure[]{
		\centering
		\includegraphics[scale=0.33]{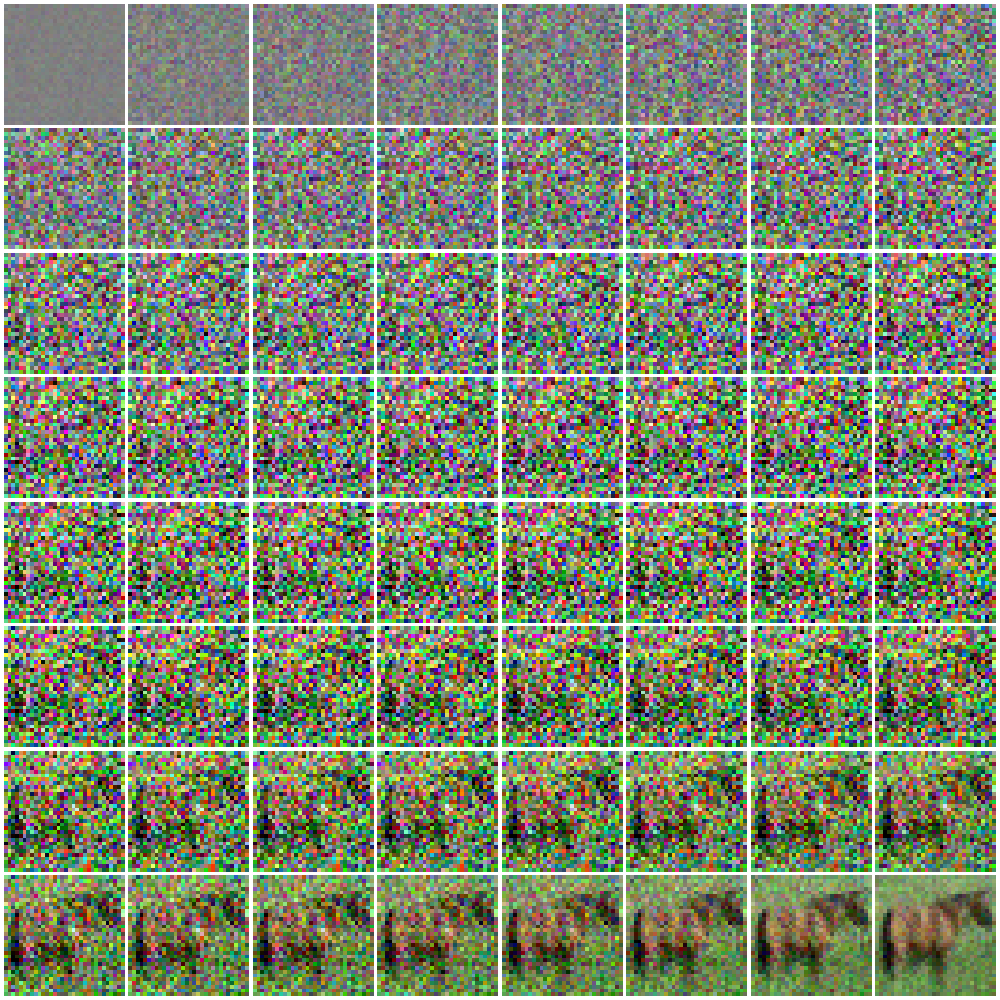} }
	\vspace{-5pt}
	\caption{Sample evolution from $0 \to 1$ : (Left) $\hat{\bm x}(t;{\bm x}(t))\doteq \sum_s\bm{y}^{(s)} w(\bm{y}^{(s)}|t;\bm{x}(t))$ and (Right) $\bm{x}(t)$. $\beta=1$} 
	\label{fig:sample_evl_1} 
\end{figure}

\begin{figure}[h!]
	\subfigure[]{
		\centering
		\includegraphics[scale=0.33]{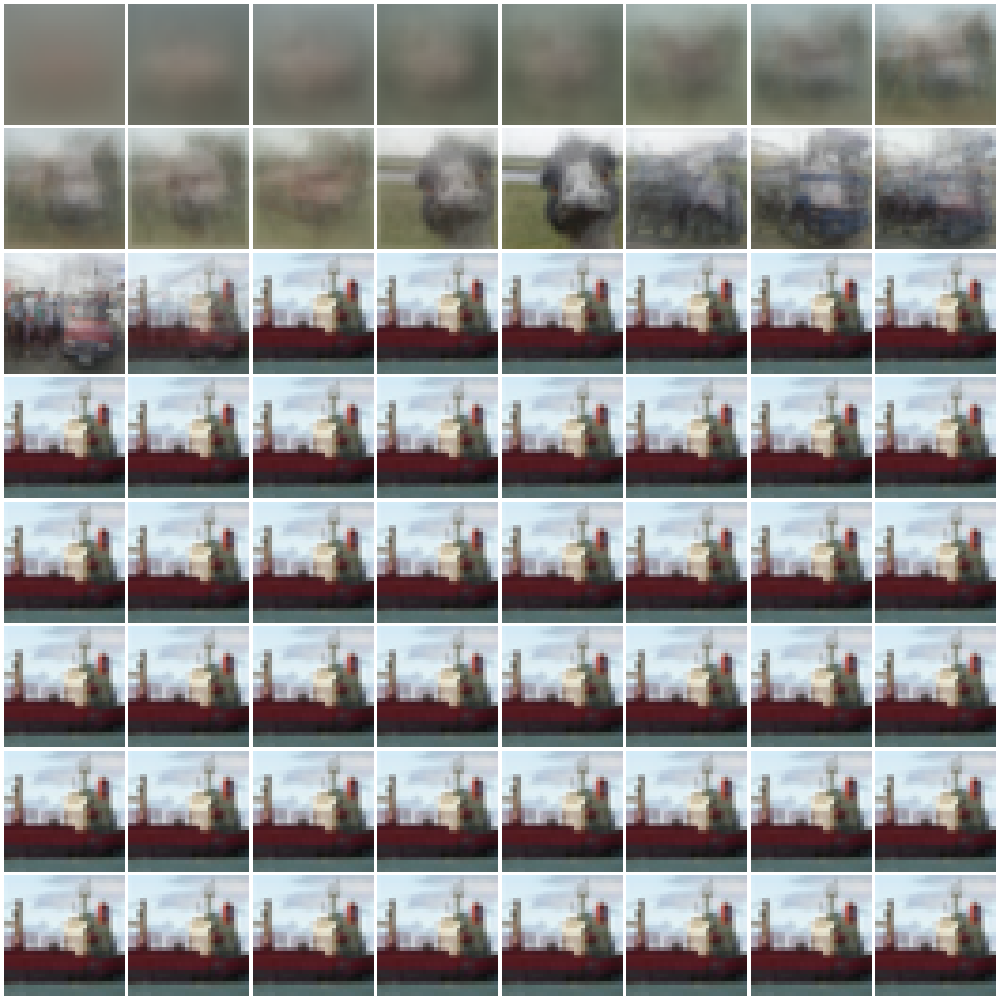} }
	\subfigure[]{
		\centering
		\includegraphics[scale=0.33]{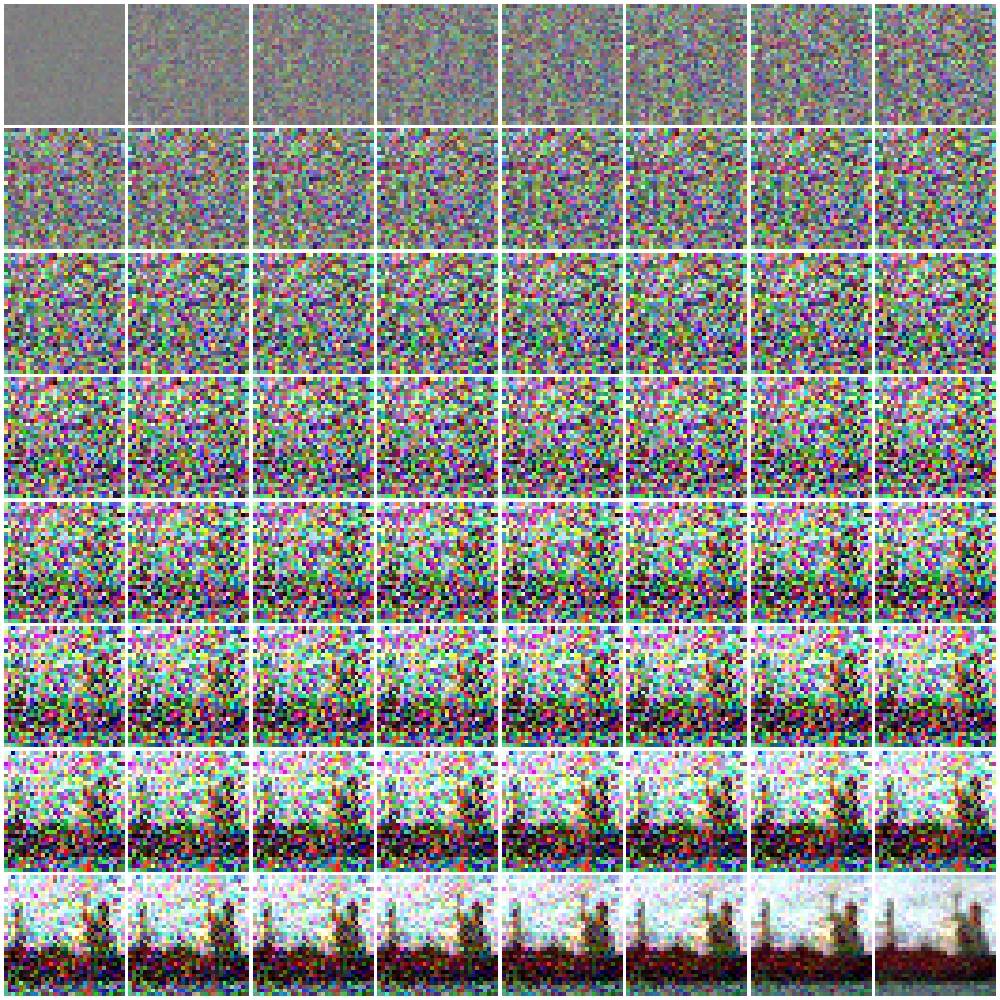} }
	\vspace{-5pt}
	\caption{Sample evolution from $0 \to 1$ : (Left) $\hat{\bm x}(t;{\bm x}(t))\doteq \sum_s\bm{y}^{(s)} w(\bm{y}^{(s)}|t;\bm{x}(t))$ and (Right) $\bm{x}(t)$. $\beta=10$} 
	\label{fig:sample_evl_10} 
\end{figure}

\begin{figure}[h!]
	\subfigure[]{
		\centering
		\includegraphics[scale=0.33]{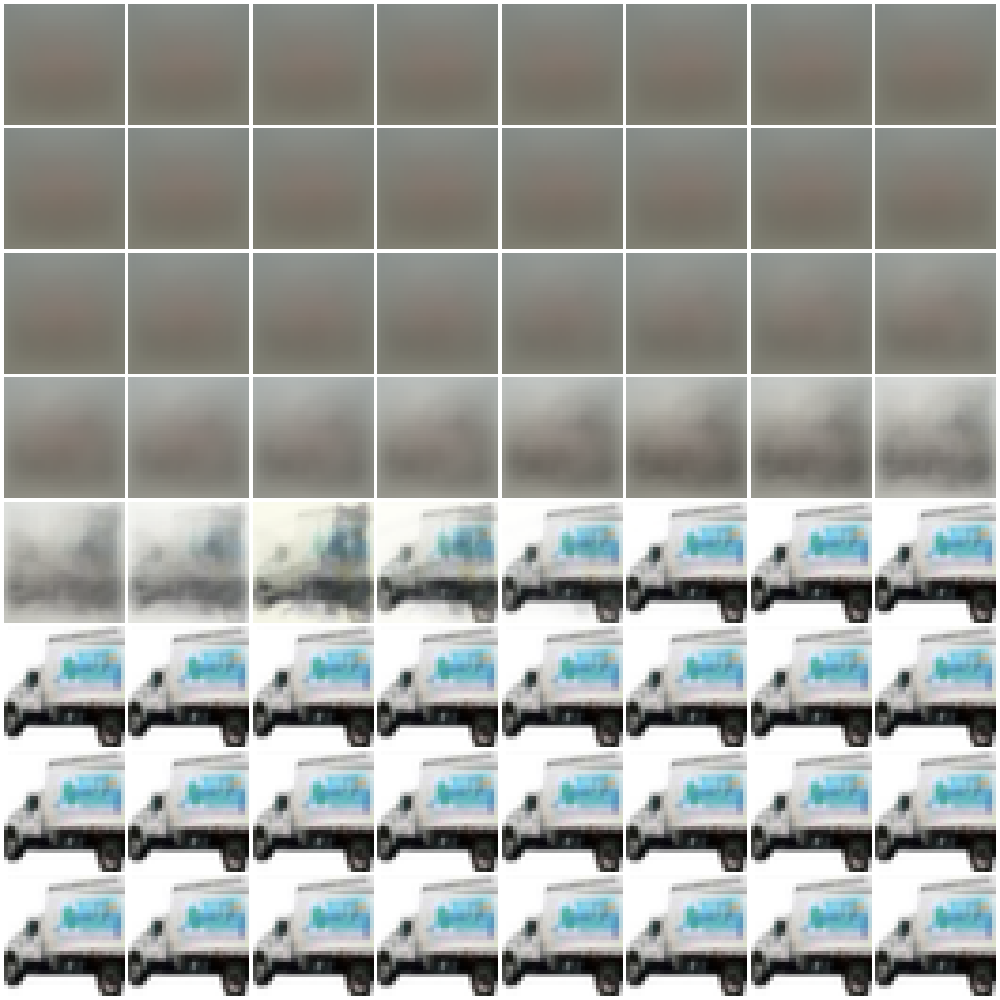} }
	\subfigure[]{
		\centering
		\includegraphics[scale=0.33]{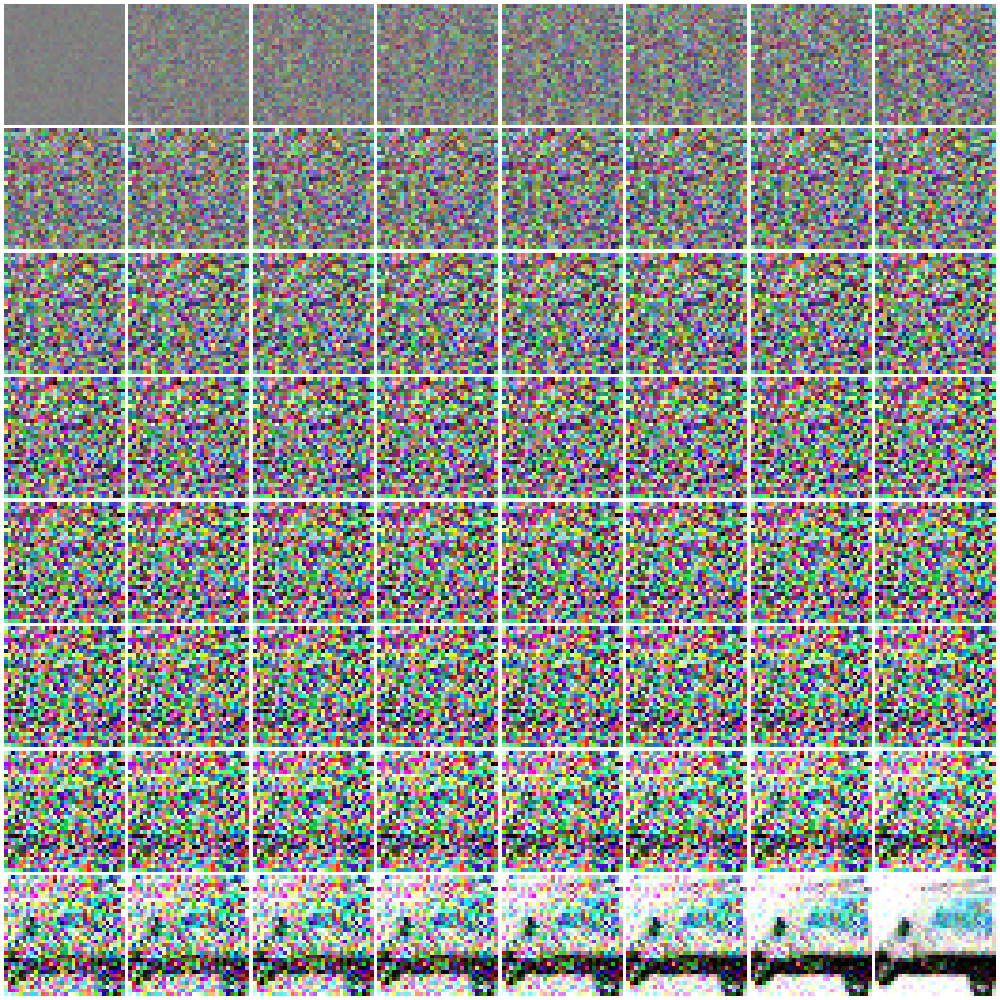} }
	\vspace{-5pt}
	\caption{Sample evolution from $0 \to 1$ : (Left) $\hat{\bm x}(t;{\bm x}(t))\doteq \sum_s\bm{y}^{(s)} w(\bm{y}^{(s)}|t;\bm{x}(t))$ and (Right) $\bm{x}(t)$. $\beta=100$} 
	\label{fig:sample_evl_100} 
\end{figure}

\clearpage


\bibliographystyle{unsrt}
\bibliography{bib/DiffusionModels.bib,bib/StochasticOptimalControl.bib, bib/Hamidreza,bib/MCMC, bib/BoltzmannGenerators}

\end{document}